\documentclass[10pt,twocolumn,letterpaper]{article}

\usepackage[accsupp]{axessibility} 
\usepackage[pagenumbers]{cvpr} 

\usepackage{graphicx}
\usepackage{amsmath}
\usepackage{amssymb}
\usepackage{booktabs}
\usepackage{multirow}
\usepackage[table]{xcolor}
\usepackage{amsmath}

\DeclareMathOperator*{\argmin}{arg\,min}

\usepackage[pagebackref,breaklinks,colorlinks]{hyperref}

\usepackage[capitalize]{cleveref}
\crefname{section}{Sec.}{Secs.}
\Crefname{section}{Section}{Sections}
\Crefname{table}{Table}{Tables}
\crefname{table}{Tab.}{Tabs.}

\def\cvprPaperID{4345} 
\def\confName{CVPR}
\def\confYear{2023}

\begin{document}

\title{Human Pose as Compositional Tokens}

\author{Zigang Geng$^{1,3}$ , Chunyu Wang$^{3}\thanks{Equal Advising}$ , Yixuan Wei$^{2,3}$, Ze Liu$^{1,3}$, Houqiang Li$^{1}$, Han Hu$^{3}\footnotemark[1]$\\
$^{1}$University of Science and Technology of China\ \ \ $^{2}$Tsinghua University\ \ \ 
$^{3}$Microsoft Research Asia\\
{\tt\small \url{https://sites.google.com/view/pctpose}}
}

\maketitle

\begin{abstract}
Human pose is typically represented by a coordinate vector of body joints or their heatmap embeddings. While easy for data processing, unrealistic pose estimates are admitted due to the lack of dependency modeling between the body joints.
In this paper, we present a structured representation, named Pose as Compositional Tokens (PCT), to explore the joint dependency. It represents a pose by $M$ discrete tokens with each characterizing a sub-structure with several interdependent joints (see \Cref{fig:token_substructure_vis}). The compositional design enables it to achieve a small reconstruction error at a low cost.
Then we cast pose estimation as a classification task. In particular, we learn a classifier to predict the categories of the $M$ tokens from an image. A pre-learned decoder network is used to recover the pose from the tokens without further post-processing. We show that it achieves better or comparable pose estimation results as the existing methods in general scenarios, yet continues to work well when occlusion occurs, which is ubiquitous in practice. The code and models are publicly available at \url{https://github.com/Gengzigang/PCT}.
\end{abstract}

\section{Introduction}
\label{sec:intro}

Human pose estimation is a fundamental task in computer vision which aims to estimate the positions of body joints from images. The recent progress has focused on network structures~\cite{sun2019hrnet, wang20hrnet,xu2022vitpose}, training methods~\cite{xie2021empirical,schmidtke2021unsupervised,iqbal2020weakly}, and fusion strategies~\cite{pavllo20193d,cheng2021graph,cheng2022dual,saini2022airpose,yi2022human,von2018recovering}, which have notably advanced the accuracy on public datasets. However, it remains an open problem in challenging scenarios, \eg, in the presence of occlusion, which hinders its application in practice.



\begin{figure*}
	\footnotesize
	\centering
\includegraphics[height=0.146\linewidth, trim={120 100 110 60},clip]{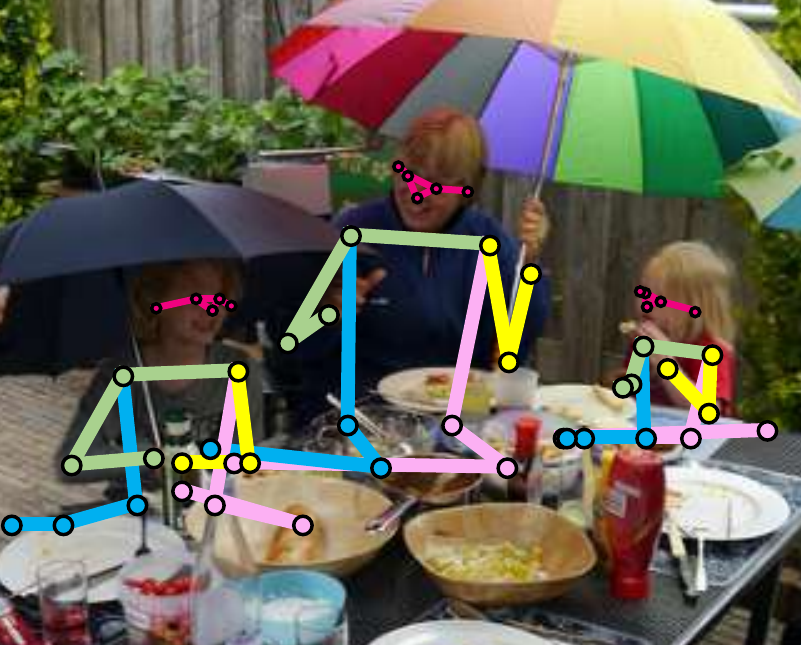}~
\includegraphics[height=0.146\linewidth, trim={40 15 60 83},clip]{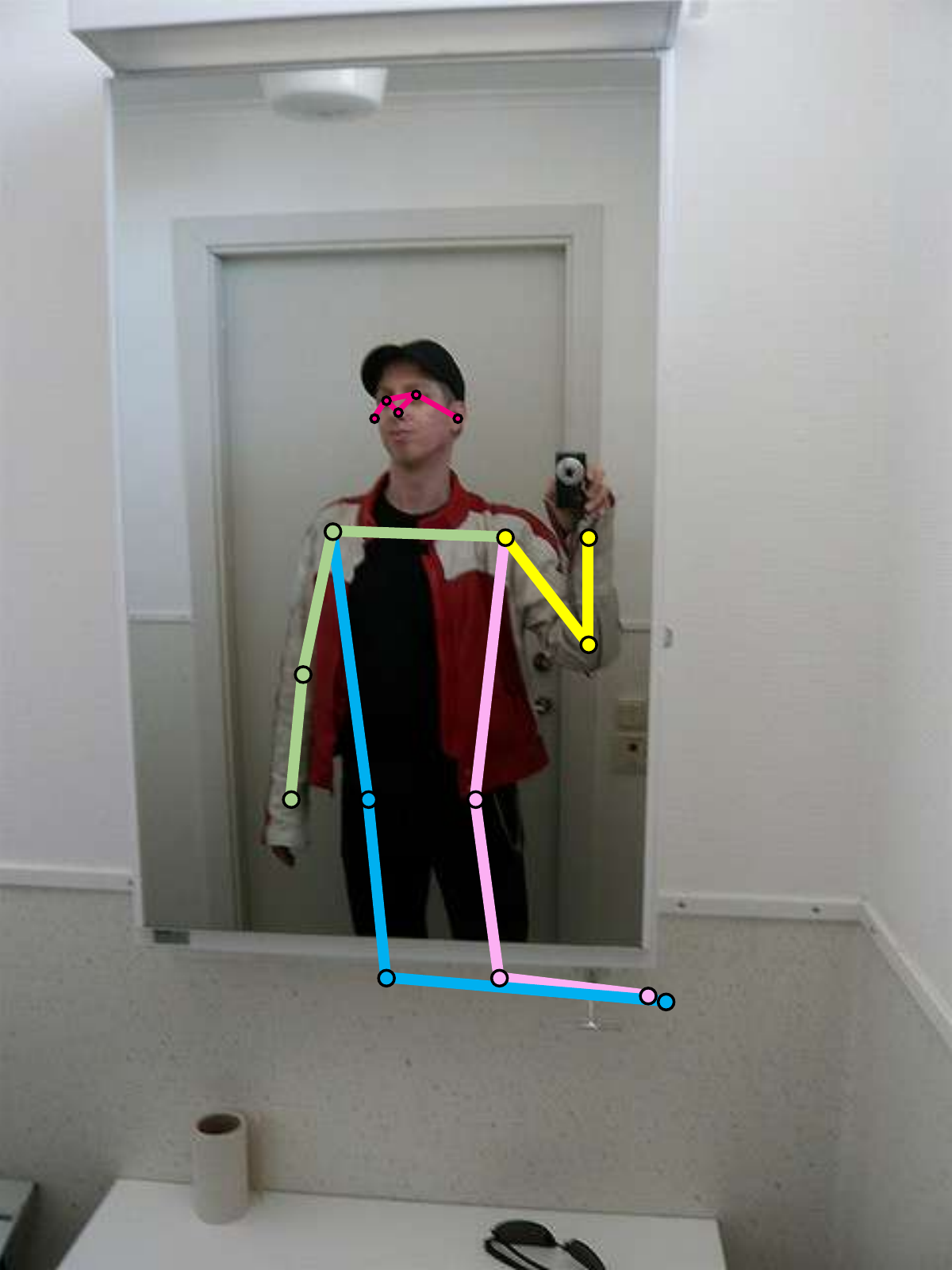}~
\includegraphics[height=0.146\linewidth, trim={70 20 0 20},clip]{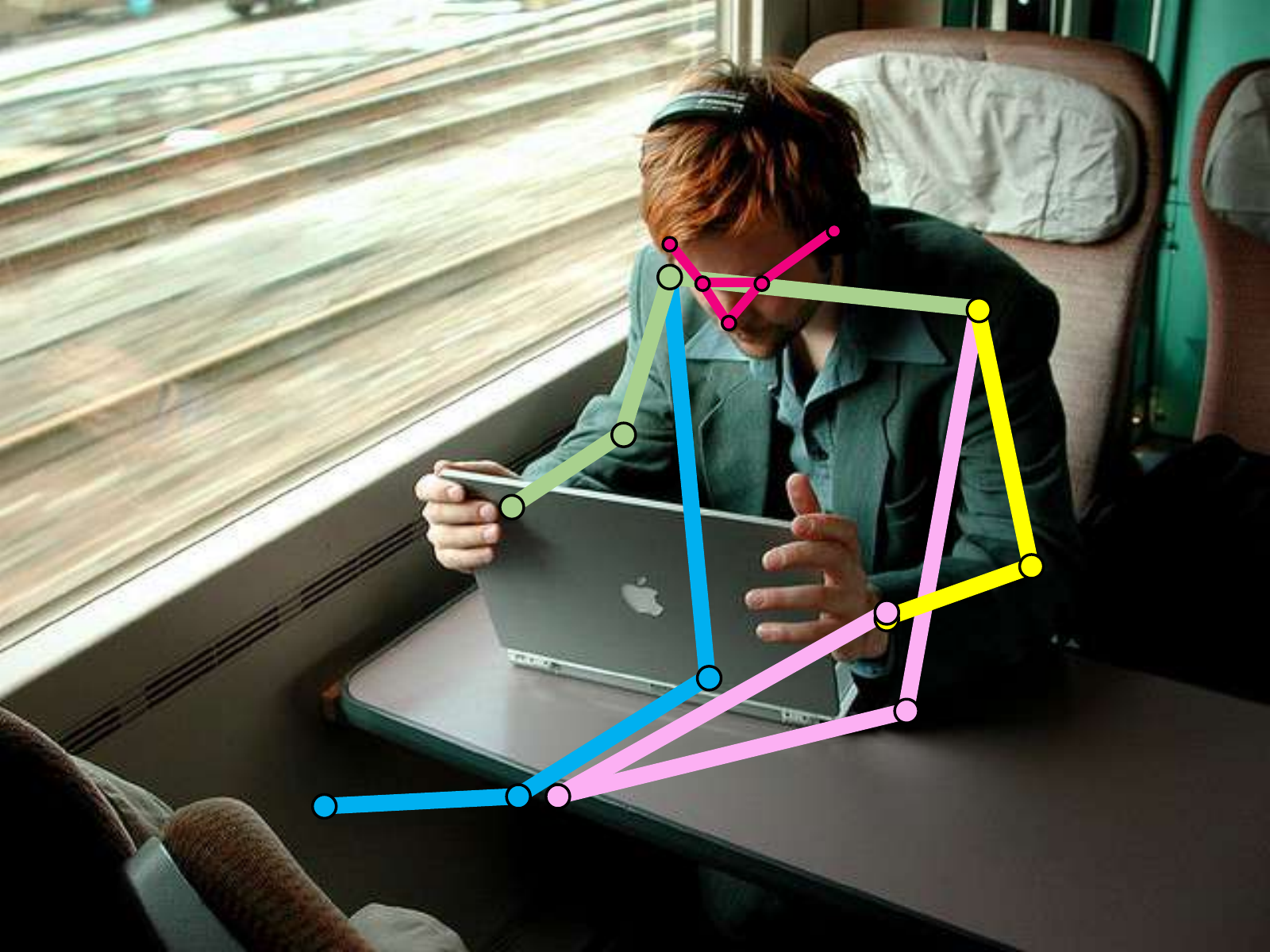}~
\includegraphics[height=0.146\linewidth, trim={90 15 0 5},clip]{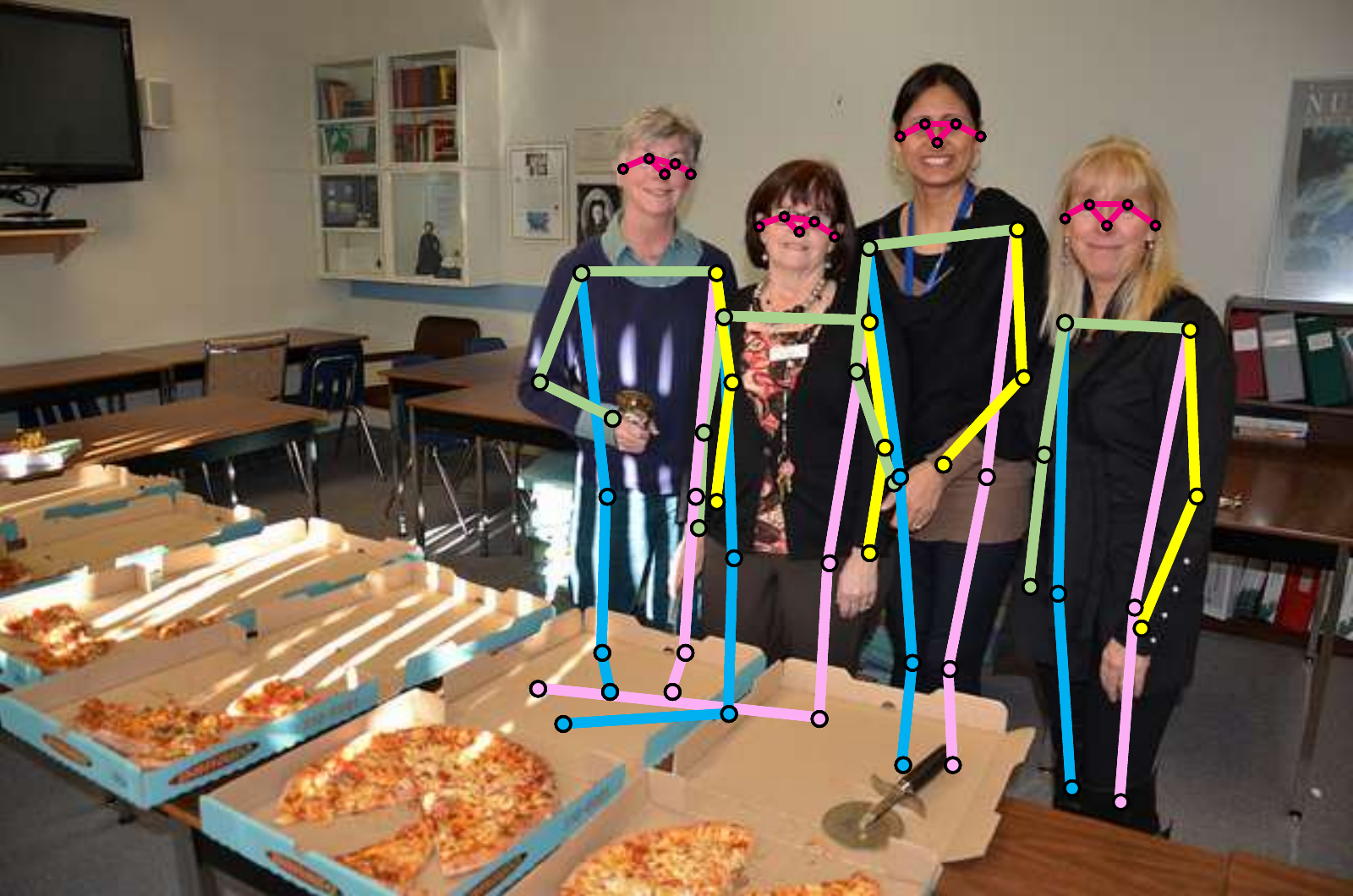}~
\includegraphics[height=0.146\linewidth, trim={0 170 40 20},clip]{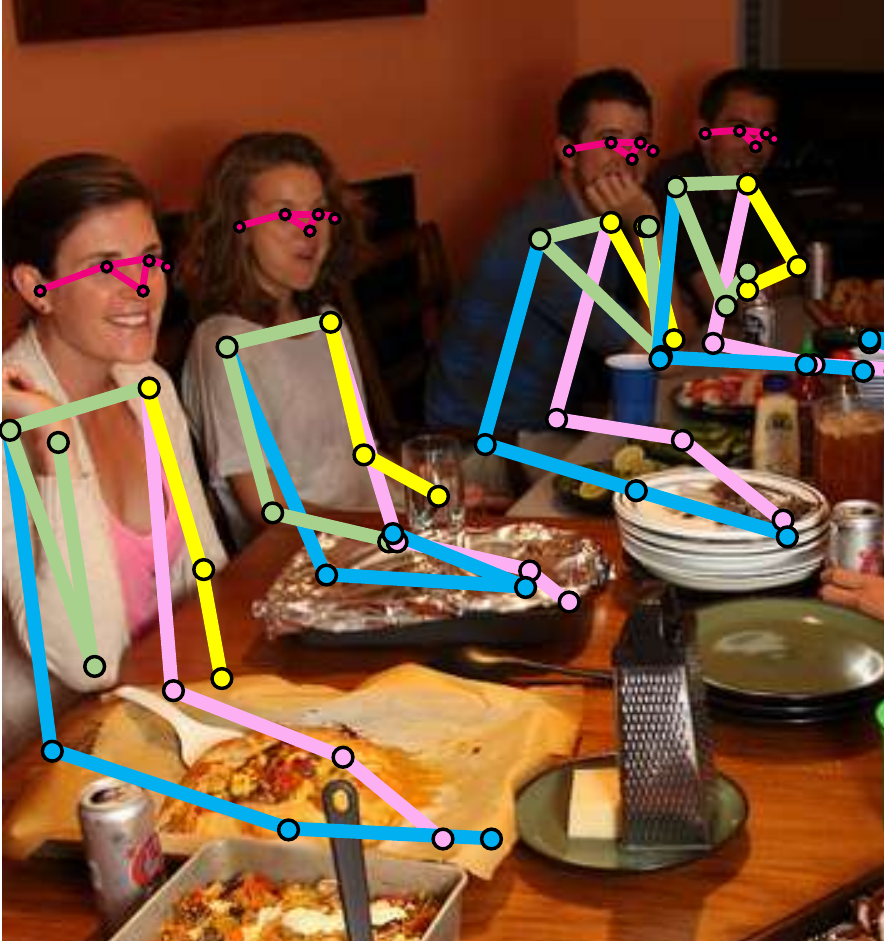}~
\includegraphics[height=0.146\linewidth, trim={120 40 80 70},clip]{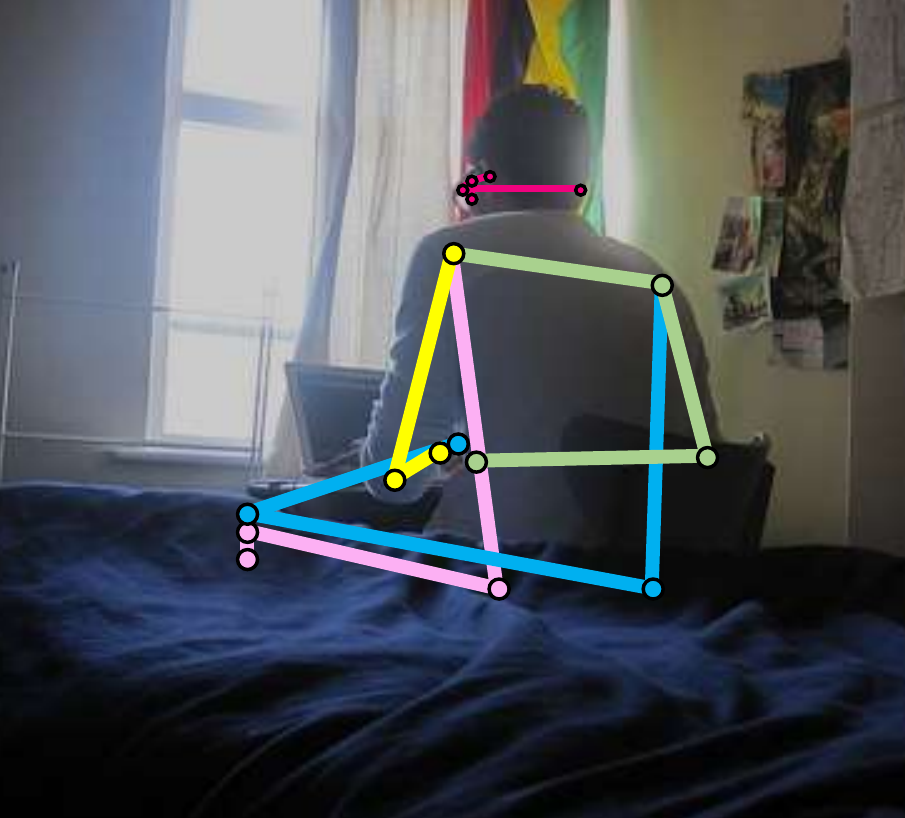}\\
\vspace{.1cm}
\includegraphics[height=0.146\linewidth, trim={120 100 110 60},clip]{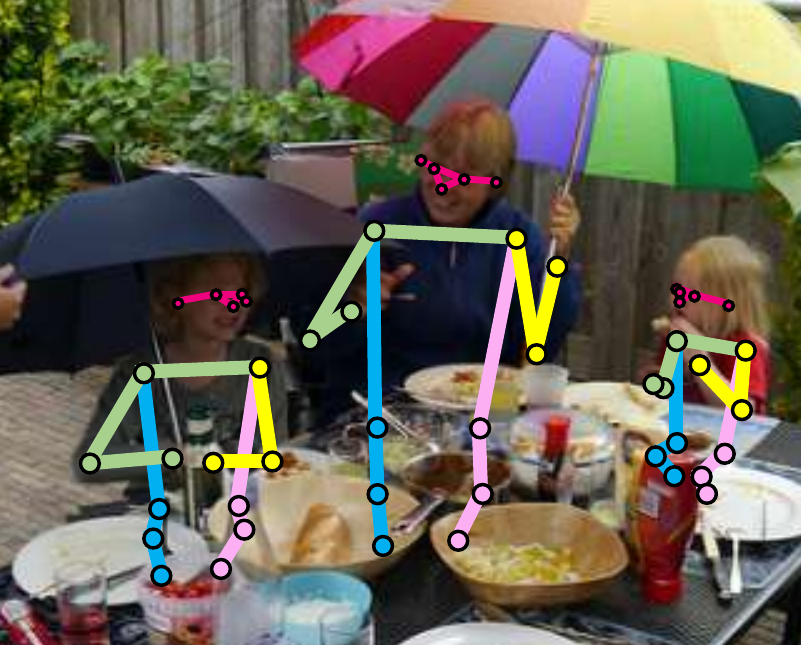}~
\includegraphics[height=0.146\linewidth, trim={40 15 60 83},clip]{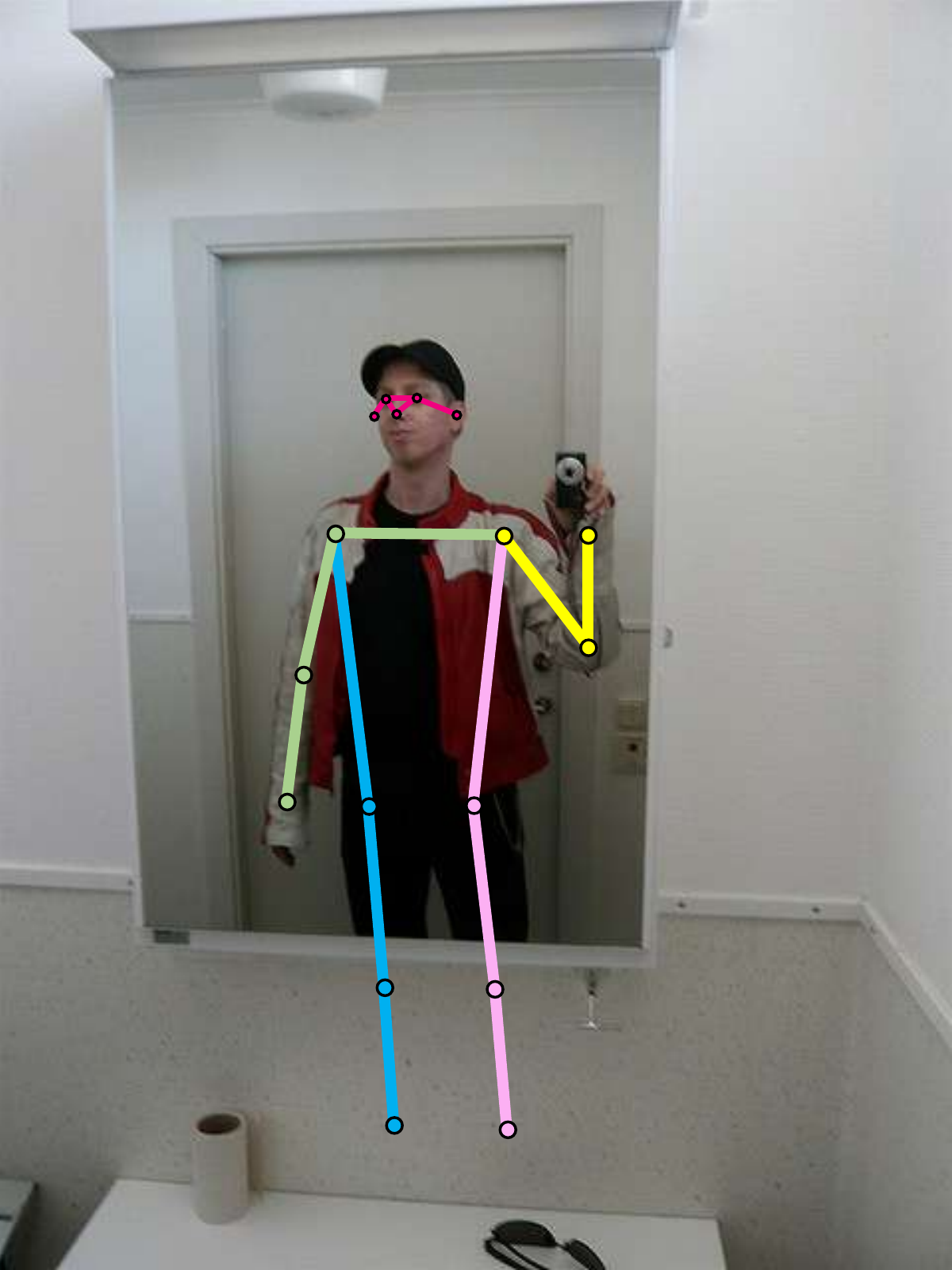}~
\includegraphics[height=0.146\linewidth, trim={70 20 0 20},clip]{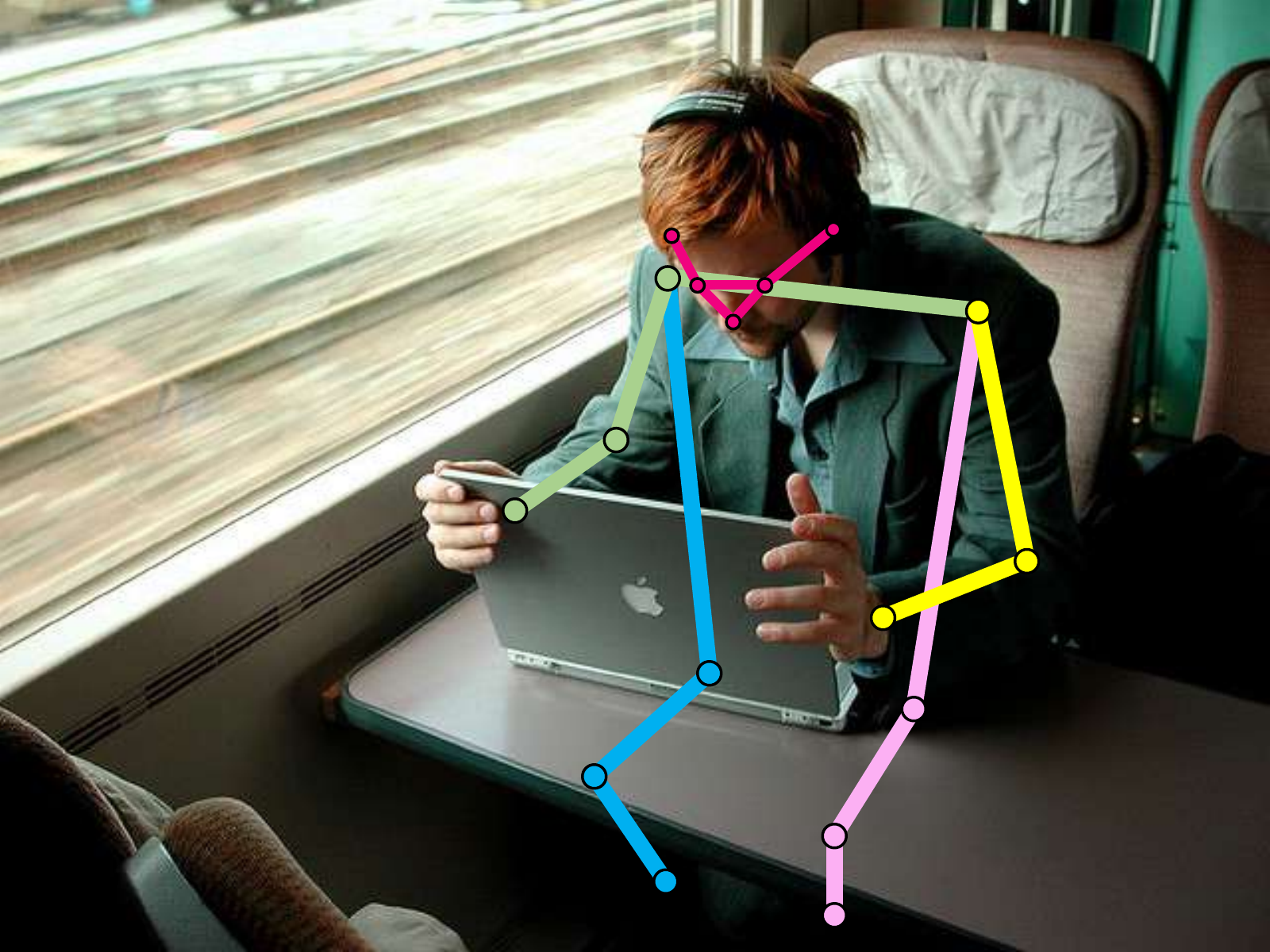}~
\includegraphics[height=0.146\linewidth, trim={90 15 0 5},clip]{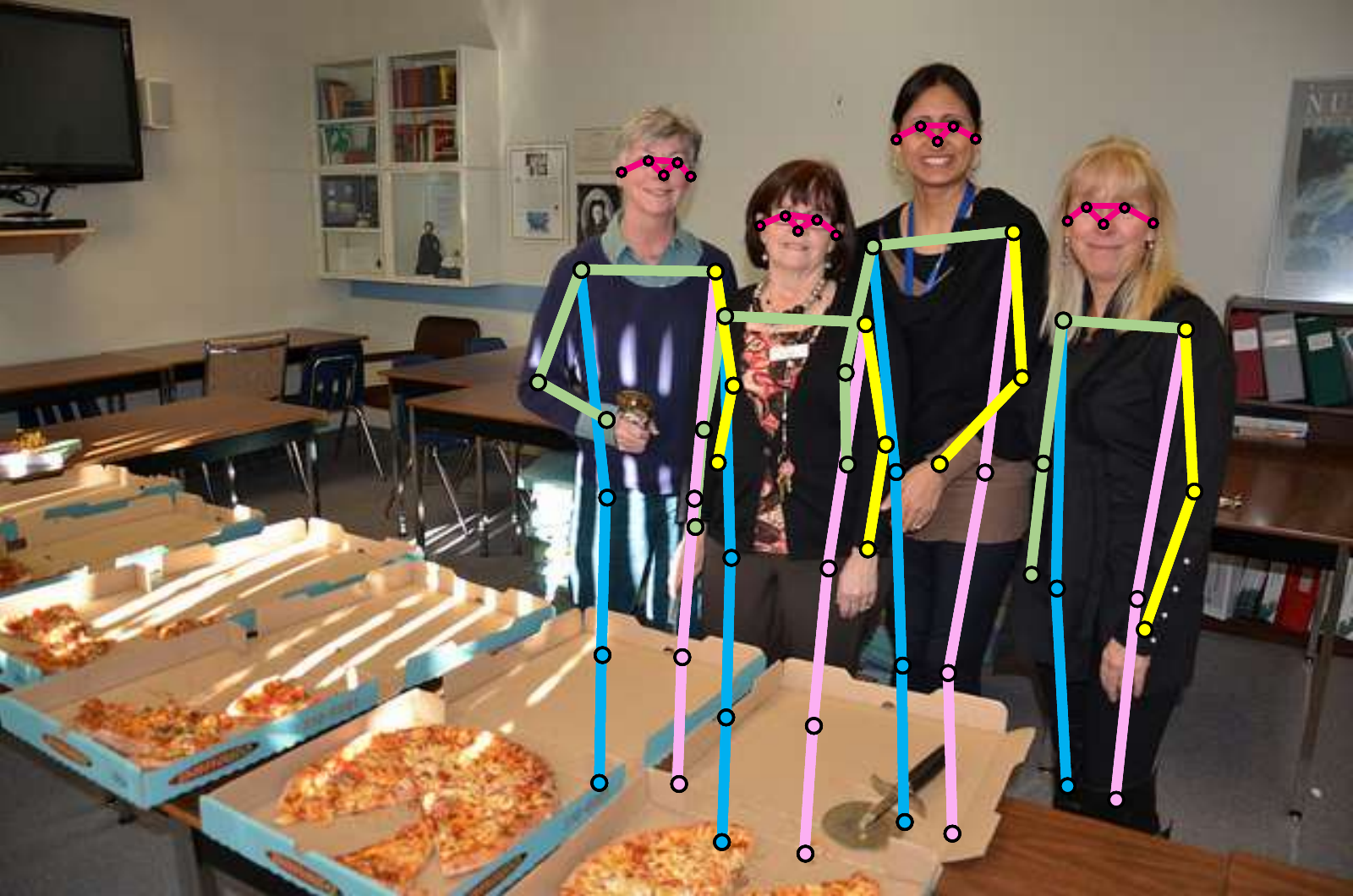}~
\includegraphics[height=0.146\linewidth, trim={0 170 40 20},clip]{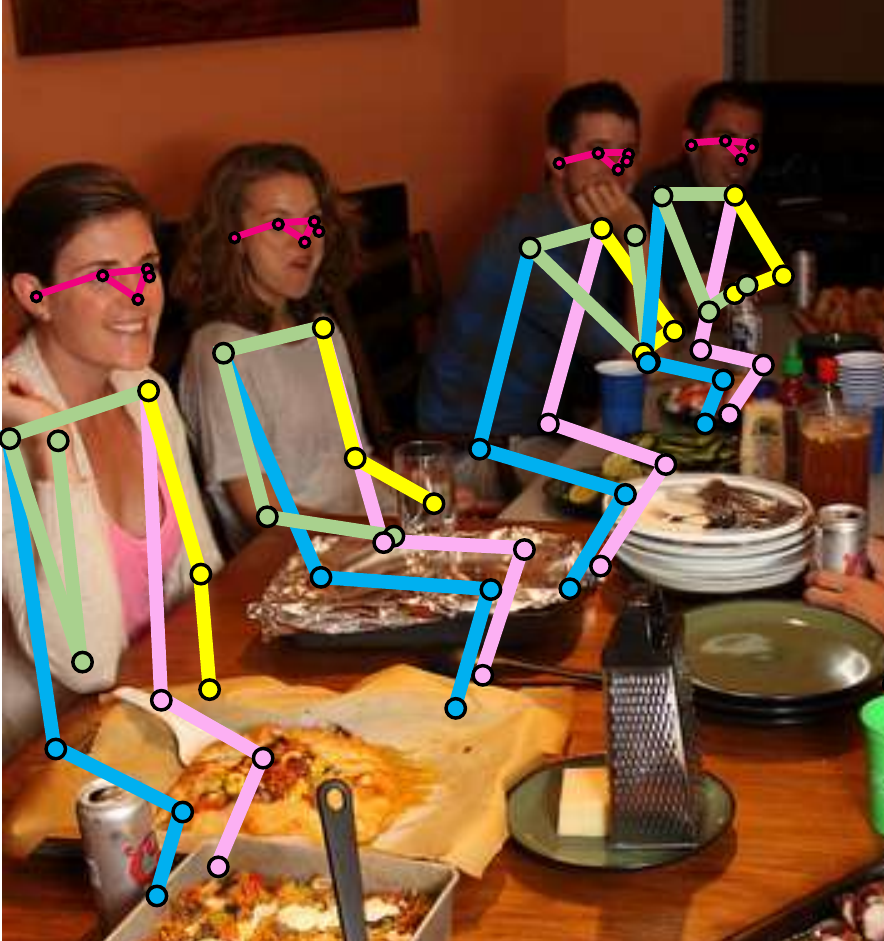}~
\includegraphics[height=0.146\linewidth, trim={120 40 80 70},clip]{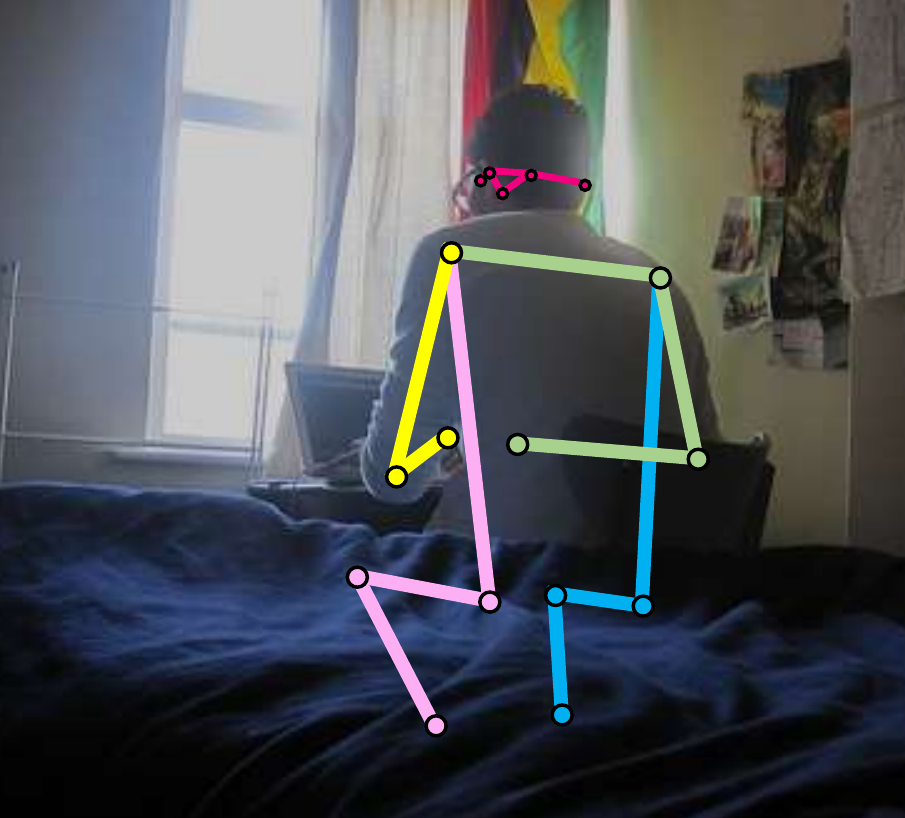}\\
	\caption{
	Heatmap-based method (\textbf{top}) v.s. our PCT method (\textbf{bottom}) in occluded scenes. PCT predicts reasonable poses even under severe occlusion.
	The images are from COCO val2017.}
	\label{fig:teaser}
\end{figure*}

Current 2/3D pose estimators usually represent a pose by a coordinate vector \cite{ToshevS14deeppose,li2021rle,Zhou19centernet,geng21dekr} or its heatmap embeddings~\cite{wei16cpm,newell16hour,Papandreou17offsetpose,sun2019hrnet,wang20hrnet,sun2018integral,tu2020voxelpose,li2021tokenpose}. In both representations, the joints are treated independently, ignoring the fact that the body joints can serve as mutual context to each other. As a result, they may get unrealistic estimates when occlusion occurs as shown in \Cref{fig:teaser} (top). However, it is interesting to note that humans can easily predict intact poses from only the visible joints and the visual features. This is probably because people are able to use context to aid recognition as evidenced by some psychology experiments~\cite{biederman1982scene,oliva2007role}.  Some works attempt to introduce a tree or graph structure~\cite{andriluka2009pictorial,felzenszwalb2005pictorial,ramanan2006learning,wang20graph} to model joint dependency. However, the hand-designed rules usually make unrealistic assumptions on the relationships, making them incapable to represent complex patterns. 

\begin{figure}[t]
	\footnotesize
	\centering
    \includegraphics[width=1.0\linewidth]{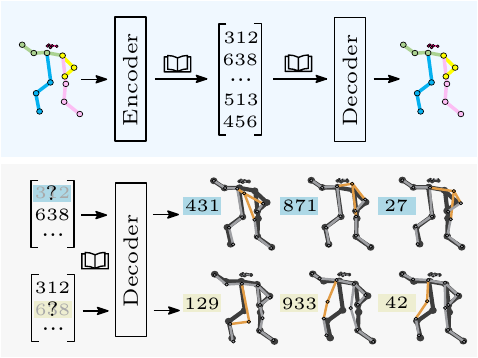}~
	\caption{Our approach represents a pose by M discrete tokens which are indices to the codebook entries (\textbf{top}).  Each token is learned to represent a sub-structure. In each row, we show that if we change the state of one token to different values, it consistently changes the same sub-structure highlighted by orange. The black poses are before changing (\textbf{bottom}).
	}
	\label{fig:token_substructure_vis}
\end{figure}

\begin{figure*}[t]
	\footnotesize
	\centering
    \includegraphics[width=1.0\linewidth, ]{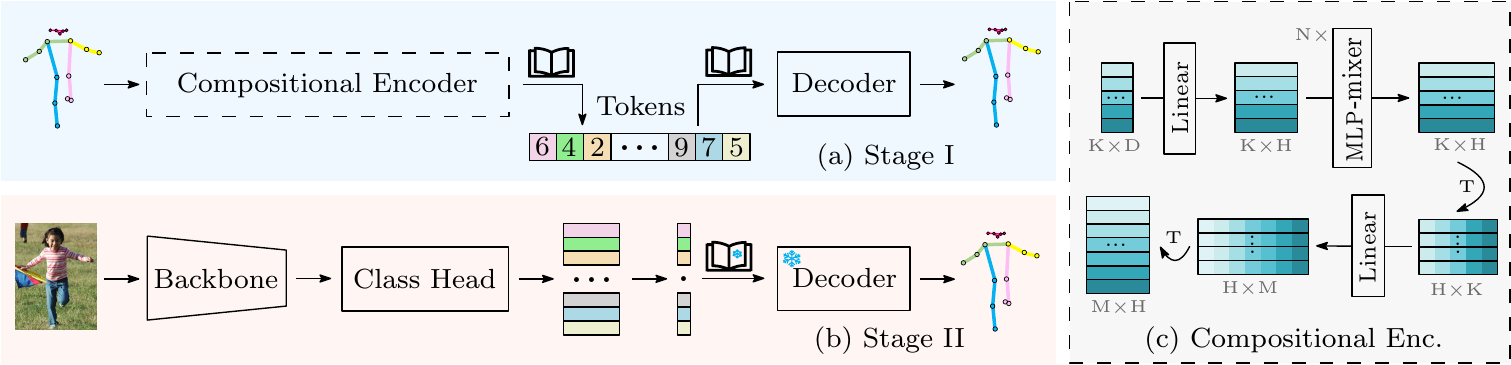}~
	\caption{ Two stages of the PCT representation (a,b)  and the structure of the compositional encoder (c). In Stage I, we learn a compositional encoder to transform a pose into $M$ tokens which are quantized by a codebook. So, a pose is represented by a set of discrete indices to the codebook. In Stage II, we cast pose estimation as a classification task by predicting the categories of the $M$ tokens, \ie the indices to the codebook entries. They will be decoded by a decoder network to obtain the final pose.
	}
	\label{fig:framework}
\end{figure*}

In this work, we hope to learn the dependency between the joints earlier in the representation stage without any assumptions. Our initial idea is to learn a set of prototype poses that are realistic, and represent every pose by the nearest prototype. While it can guarantee that all poses are realistic, it requires a large number of prototypes to reduce the quantization error to a reasonable level which is computationally infeasible. Instead, we propose a discrete representation, named pose as compositional tokens (PCT).
Figure \ref{fig:framework} shows the two stages of the representation. In Stage I, we learn a compositional encoder to transform a pose into $M$ token features, with each encoding a sub-structure of the pose. See Figure \ref{fig:token_substructure_vis} for some examples. Then the tokens are quantized by a shared codebook. So, a pose is simply represented by $M$ discrete indices. The space represented by the codebook is sufficiently large to represent all poses accurately. We jointly learn the encoder, the codebook, and the decoder by minimizing a reconstruction error.   

In Stage II, we cast human pose estimation as a classification task. Given an image, we predict the categories of the $M$ tokens, from which the pose is recovered by the decoder network. The PCT representation has several advantages.
First, the dependency between the joints is modeled by the tokens, which helps to reduce the chance of getting unrealistic pose estimates. In particular, we see evidence that it has the potential to obtain reasonable estimates even when a large portion of the body is occluded. See Figure \ref{fig:teaser} (bottom) for some examples. Second, it does not require any expensive post-processing modules such as UDP~\cite{huang20udp} which is required by the heatmap representation to reduce the quantization errors. Third, it provides a unified representation for 2D and 3D poses. In addition, the discrete representation potentially facilitates its interactions with other discrete modalities such as text and speech. But this is not the focus of this work.

We extensively evaluate our approach in 2D human pose estimation on five benchmark datasets. It gets better or comparable accuracy as the state-of-the-art methods on all of them. But more importantly, it achieves significantly better results when evaluated only on the occluded joints, validating the advantages of its dependency modeling capability. We also present the results in 3D pose estimation on the H36M dataset on which it achieves comparable accuracy as the state-of-the-art methods using a simple architecture. The results demonstrate that it has wide applicability. 

\section{Related works}
\label{sec:related}
In this section, we first briefly discuss the widely used pose representations. Then we discuss the methods that explore joint dependencies.

\subsection{Pose representations}

\noindent\textbf{Coordinates.}
Early works~\cite{ToshevS14deeppose, Belagiannis15reg, Carreira16IEF, Zhou19centernet, nie19ssm, tian19directpose, mao21fcpose} propose to directly regress the coordinates of body joints from images. While efficient, the accuracy is worse than the heatmap-based methods because it is challenging to learn the highly non-linear mapping. Some works~\cite{wei20pointanchor, geng21dekr} propose to improve them by focusing on local features around the joints. Residual Log-likelihood Estimation~\cite{li2021rle} proposes a novel regression paradigm to capture the underlying output distribution. MDN~\cite{Varamesh20MDN} introduces mixture density network for regression. Recently, transformer~\cite{vaswani17transformer} brings notable improvement~\cite{li2021prtr, shi22petr, mao2022poseur} due to its ability to capture long-range information.
\vspace{0.2em}

\noindent\textbf{Heatmaps.}
The heatmap representation~\cite{li16voting, cao2017realtime, yang17pyramid, fang17RMPE, xiao18simpleb, peng18aug, newell17ae, Sekii18, li19multistage, Artacho20unipose, luo21swhar, wang22litepose} has been dominant since its introduction~\cite{tompson2014joint, bulat16cpmr, wei16cpm} because of its strong localization and generalization ability. Many follow-up works have been devoted to continuously improving them, including proposing powerful networks~\cite{newell16hour, he17maskrcnn, chen18cpn, sun2019hrnet, cai20rsn, cheng2019higher} to estimate the heatmaps more accurately, introducing the attention operator to the models~\cite{su19attn, yuan21hrformer, yang2021transpose, li2021tokenpose}, reducing the quantization errors~\cite{huang20udp, zhang20dark}, 
fusion with the coordinate prediction-based methods~\cite{fan15dual, Papandreou17offsetpose, sun2018integral, gu21biasreg}, refining the results~\cite{sun17normrefine, fieraru18refine, moon19refine, wang20graph}, leveraging other tasks~\cite{nie18parsing, Kocabas18fast, papandreou18personlab}, and leveraging large unlabeled datasets~\cite{xie2021empirical,kim2022pose}. However, the heatmap representation suffers from quantization errors caused by the down-sampling operations in neural networks. Besides, the joint dependency is not modeled by the heatmaps. 


\vspace{0.2em}
\noindent\textbf{Discrete bins.}
Recent works~\cite{li2021simcc,lu2022unifiedio, chen2022pix2seqv2} propose to divide each pixel into several bins, allowing sub-pixel localization accuracy. The horizontal and vertical coordinates of each joint are separately quantized into discrete classes. Similar to our work, they also cast human pose estimation as a classification task. However, each coordinate of the pose is treated independently which differs from our structured representation.

\subsection{Modeling joint dependency}
Since the human body has an articulated structure, there are many works trying to model joint dependency to help resolve low-level ambiguities. However, most of them focus on the modeling aspect rather than representation which is the focus of this work.

\vspace{0.2em}
\noindent\textbf{Pictorial structures.}
Some works~\cite{andriluka2009pictorial,felzenszwalb2005pictorial,ramanan2006learning, yang11mop, Pishchulin13poseletcond} propose to use the deformable model where the relationship between body joints is explicitly considered based on anatomy priors (\eg limb lengths). However, they have three disadvantages. First, they usually make strong assumptions on the relationships, \eg 
Gaussian distribution on the offsets between two joints, making them incapable to represent complex patterns. Second, they still require that the body joints can be independently detected from images first, and based on that they use the dependency priors to obtain the most plausible configuration. However, the first step is already very difficult in cluttered scenes with serious occlusions. Finally, they cannot be trained end-to-end with the deep networks with an exception ~\cite{tompson2014joint} that needs to relax the formulation.
\vspace{0.2em}

\noindent\textbf{Implicit modeling.}
The recent deep learning-based methods~\cite{chu16structure, yang16deepmop, zhang19graphcontext, wang20graph, yang21gcn, qiu29peekocc} implicitly model the dependency by propagating the visual features between the joints. For example, Chu \etal ~\cite{chu16structure} introduce geometrical transform kernels to fuse the features of different channels which are believed to characterize different joints. Wang \etal ~\cite{wang20graph} use Graph Convolutional Network to refine pose estimates which are obtained by the heatmap-based methods first. In addition, Chen \etal~\cite{chen17adversial} propose to learn a pose discriminator to exclude non-realistic pose estimates and push the predictor to learn poses with reasonable structures. Li \etal~\cite{li2021tokenpose} explicitly learn a type embedding for each joint and apply the transformer to model the relationships among the joints. But from the aspect of representation, they still treat each joint independently and predict the heatmap for each joint.

Our PCT representation differs from the previous methods in three aspects. First, the joint dependency is encoded earlier in the representations by the tokens (changing the state of a token changes the corresponding sub-structure rather than a single joint). In contrast, the other three representations treat each joint independently.  Second, the sub-structures are automatically learned from training data without making any unrealistic assumptions. We empirically show that it has a stronger capability to resolve ambiguities caused by occlusion in a variety of situations. Third, the joint dependency is explicitly imposed rather than by implicit feature propagation. The latter method still allows unrealistic pose estimates in challenging situations.


\section{Pose as Compositional Tokens}
In Section \ref{sec:pct_codebook}, we describe how to learn the codebook and the encoder/decoder networks. Section \ref{sec:pct_downstream} explains how it is used in the human pose estimation task.


\subsection{Learning compositional tokens}
\label{sec:pct_codebook}
We represent a raw pose as $\mathbf{G} \in \mathbb{R}^{K \times D}$ where $K$ is the number of body joints and $D$ is the dimension of each joint, where $D=2$ for 2D pose, and $D=3$ for 3D pose, respectively. We learn a compositional encoder $f_e{(\cdot)}$ to transform a pose into $M$ token features:
\begin{equation}
    \mathbf{T}=(\mathbf{t}_1, \mathbf{t}_2, \cdots, \mathbf{t}_M)=f_e{(\mathbf{G})},
\end{equation}
where each token feature $\mathbf{t}_i \in \mathbb{R}^H$ approximately corresponds to a sub-structure of the pose which involves a few interdependent joints. Figure \ref{fig:token_substructure_vis} shows some of the learned examples. Note that the representation has lots of redundancy because different tokens may have overlapping joints. The redundancy makes it robust to occlusions of individual parts.  

Figure \ref{fig:framework} (c) shows the network structure of the encoder. The position of each body joint is first fed to a linear projection layer to increase the feature dimension. Then the features are fed to a series of MLP-Mixer~\cite{TolstikhinHKBZU21mlpmixer} blocks to deeply fuse the features of different joints. Finally, we extract $M$ token features by applying a linear projection to the features across all of the joints.

Similar to~\cite{vqvae}, we define a latent embedding space by a codebook $\mathbf{C}=(\mathbf{c}_1, \cdots, \mathbf{c}_V)^{\operatorname{T}} \in \mathbb{R}^{V \times N}$ where $V$ is the number of codebook entries. We quantize each token $\mathbf{t}_i$ by the nearest neighbor look-up using the embedding space as shown in the following equation:
\begin{equation}
q(\mathbf{t}_i=v|\mathbf{G}) = \left\{
\begin{aligned}
1 & \quad \text{if} \quad v=\argmin_j{\|\mathbf{t}_i-\mathbf{c}_j\|_2}\\
0 & \quad \text{otherwise}
\end{aligned}
\right.
\end{equation}
Note that all tokens share the same embedding space $\mathbf{C}$ which simplifies training.

We abuse $q({\mathbf{t}_i})$ to represent the index to the corresponding codebook entry.  Then the quantized tokens $(\mathbf{c}_{q({\mathbf{t}_1})}, \mathbf{c}_{q({\mathbf{t}_2})}, \cdots, \mathbf{c}_{q({\mathbf{t}_M})})$ will be fed to the decoder network to recover the original pose:
\begin{equation}
    \mathbf{\hat{G}}=f_d{(\mathbf{c}_{q({\mathbf{t}_1})}, \mathbf{c}_{q({\mathbf{t}_2})}, \cdots, \mathbf{c}_{q({\mathbf{t}_M})})}
\end{equation}
The network structure is similar to the encoder network in the reverse order except that we use a shallower MLP-Mixer network with only one block.

The encoder network, the codebook, and the decoder network are jointly learned by minimizing the following loss over the training dataset:
\begin{equation}
    \ell_{pct}={\text{smooth}_{L_1}(\hat{\mathbf{G}}, \mathbf{G}}) + \beta \sum\limits_{i = 1}^{M} \|\mathbf{t}_i-\operatorname{sg}[\mathbf{c}_{q({\mathbf{t}_i})}]\|_2^2,
\end{equation}
where, $\operatorname{sg}$ denotes stopping gradient, $\beta$ is a hyperparameter. 

We follow the optimization strategy used in~\cite{vqvae} to handle the broken gradients issue in the quantization step and the codebook is updated using the exponential moving average of previous token features. In our implementation, we have two designs that improve the results. First,
inspired by~\cite{he2022masked,xie2022simmim}, we randomly mask some joints and require the model to reconstruct them. Second, we concatenate the image features around the joints with the positional features to enhance its discrimination ability.

\paragraph{Discussion.} We try to explain why PCT learns tokens that correspond to meaningful sub-structures of poses. At one extreme, if each token corresponds to a single joint, then we need $w \times h$ (\ie $65536$ for an image of size $256 \times 256$) codebook entries to achieve a small quantization error. But we only use $1024$ entries in our experiments which is much smaller. This drives the model to learn larger structures than individual joints to improve the efficiency of the codebook. At another extreme, if we let a token correspond to an intact pose, then we only need one token instead of $M$ tokens. But in the worst case, it requires $(wh)^K$ codebook entries in order to quantize the poses with a small error. In contrast, our method drives the model to divide a pose into multiple basic sub-structures whose possible configurations can be described by a shared set.

\paragraph{Relation to VQ-VAE~\cite{vqvae}.} The PCT representation is inspired by VQ-VAE. The main difference is that VQ-VAE treats well-defined regular data, \eg image patches with the resolution of $16 \times 16$, as tokens. However, for human poses, we require PCT to automatically learn meaningful sub-structures as tokens, which is realized by the compositional encoder as well as the codebook sharing scheme. Besides, the network structures of the encoder and decoder are particularly designed for human poses, different from VQ-VAE. 

\subsection{Human Pose Estimation}
\label{sec:pct_downstream}
With the learned codebook and the decoder, we cast human pose estimation as a classification task. As shown in Figure \ref{fig:framework}, given a cropped input image $\mathbf{I}$, we simply predict the categories of the $M$ tokens, which are fed to the decoder to recover the pose. We use backbone for extracting image features $\mathbf{X}$ and design the following classification head.

\paragraph{Classification head.}
We first use two basic residual convolution blocks~\cite{He15resnet} to modulate the backbone features. Then, we flatten the features and change their dimension by a linear projection layer:
\begin{equation}
   \mathbf{X}_f = \mathcal{L}(\operatorname{Flatten}(\mathcal{C}(\mathbf{X}))), 
  \label{eq:vqvaeloss}
\end{equation}
where $\mathcal{C}$ and $\mathcal{L}$ represent the feature modulator and the linear projection respectively. We reshape the one-dimensional output feature into a matrix $\mathbf{X}_f \in \mathbb{R}^{M \times N}$, use four MLP-Mixer blocks~\cite{TolstikhinHKBZU21mlpmixer} to process the features, and output the logits of token classification:
\begin{equation}
   \mathbf{\hat{L}} = \mathcal{M}(\mathbf{X}_f),
  \label{eq:vqvaeloss}
\end{equation}
where $\mathbf{\hat{L}}$ has the shape of $\mathbb{R}^{M \times V}$.

\paragraph{Training.}
We use two losses to train the classification head. First, we enforce the cross entropy loss:
\begin{equation}
   \ell_{cls} = \operatorname{CE}(\mathbf{\hat{L}}, \mathbf{L}),
  \label{eq:classloss}
\end{equation}
where $\mathbf{L}$ denotes the ground-truth token classes obtained by feeding the ground-truth poses into the encoder.

We also enforce a pose reconstruction loss, which minimizes the difference between the predicted and the ground-truth poses. To allow the gradients from the decoder network to flow back to the classification head, we replace the hard inference scheme with a soft version:
\begin{equation}
  \mathbf{S} = \mathbf{\hat{L}} \times \mathbf{C},
  \label{eq:clssoft}
\end{equation}
where $\mathbf{S} \in \mathbb{R}^{M \times N}$ denotes the linearly interpolated token features. The token features $\mathbf{S}$ are then fed to the pre-learned decoder to obtain the predicted pose $\mathbf{\hat{G}}$. The complete loss function is:
\begin{equation}
   \ell_{all} = CE(\mathbf{\hat{L}}, \mathbf{L}) + \operatorname{smooth}_{L_1}(\mathbf{\hat{G}}, \mathbf{G}).
  \label{eq:classwholeloss}
\end{equation}
Note that the decoder network is not updated during training.

\section{Experiments}
We first extensively evaluate the PCT representation on five benchmark datasets in the context of 2D human pose estimation. Then we present the 3D pose estimation results and compare them to the state-of-the-art methods. Ablation studies about the main components of our method are also provided to help understand the approach.

\label{sec:experiments}
\subsection{Datasets and metrics}
\paragraph{2D pose datasets.} First, we conduct experiments on the COCO~\cite{lin14coco} and MPII~\cite{Andriluka14mpii} datasets. The COCO dataset has $150K$ labeled human instances for training, $5K$ images for validation, and $30K$ images for testing. The MPII dataset has $40K$ labeled human instances performing a variety of activities. Second, we evaluate our method on four datasets that have severe occlusions, including the test set of the CrowdPose~\cite{li2018crowdpose} dataset, the validation and test sets of the OCHuman~\cite{zhang19ochuman} dataset, and the SyncOCC~\cite{zhang2020syncocc} dataset. In CrowdPose~\cite{li2018crowdpose} and OCHuman~\cite{zhang19ochuman}, the occluded joints are manually labeled by annotators. The SyncOCC~\cite{zhang2020syncocc} dataset is a synthetic dataset generated by UnrealCV~\cite{qiu2017unrealcv} so it provides accurate locations of the occluded joints. We directly apply the model trained on the COCO dataset to the four datasets without re-training. We report the results on the occluded joints to validate the capability of the model to handle occlusion.

\paragraph{3D pose datasets.}
We conduct experiments on the Human3.6M~\cite{Ionescu14human36m} dataset which has 11 human subjects performing daily actions. We follow the practice of the previous works such as~\cite{ci2019optimizing}. In particular, five subjects (S1, S5, S6, S7, S8) are used for training, and two subjects (S9, S11) are used for testing. Since there are no labels for joint occlusion, we only 
compare our method to the state-of-the-art methods to validate the general applicability of the representation to both 2D and 3D poses.

\paragraph{Evaluation metrics.} 
We follow the standard evaluation metrics for the COCO~\cite{lin14coco}, MPII~\cite{Andriluka14mpii} and, Human3.6M~\cite{Ionescu14human36m} datasets. In particular, the OKS-based $\operatorname{AP}$ (average precision), $\operatorname{AP}^{50}$ and $\operatorname{AP}^{75}$ are reported for the COCO dataset. The $\operatorname{PCKh}$ (head-normalized probability of correct keypoint) score is used for the MPII dataset. The $\operatorname{MPJPE}$ (mean per joint position error) are used for Human3.6M. On the four occlusion datasets, we report the $\operatorname{AP}^{OC}$ based on OKS computed only on the occluded joints.

\subsection{Implementation details} 
We adopt the top-down estimation pipeline. In training, we use the GT boxes provided by the datasets. In testing, we use the detection results provided by~\cite{xiao18simpleb} for COCO, and the GT boxes for MPII and the occlusion datasets following the common practice. 

We use the Swin Transformer V2~\cite{liu2021Swin, liu2022swinv2} backbone pretrained with SimMIM~\cite{xie2022simmim} on ImageNet-1k~\cite{RussakovskyDSKS15imagenet}. It is also trained on the COCO dataset with heatmap supervision. To save computation cost, we fix the backbone and only train the classification head. We set the base learning rate, weight decay and batch size to $8e$-$4$, $0.05$ and $256$, respectively. In total, we train the head for $210$ epochs on COCO and MPII, and $50$ epochs on Human3.6M. The flip testing is used.

We use the default data augmentations provided by MMPose~\cite{mmpose2020} including random scale ($0.5$, $1.5$), random rotation ($-40^{\circ}$, $40^{\circ}$), random flip (50\%), grid dropout and color jitter (h=$0.2$, s=$0.4$, c=$0.4$, b=$0.4$). We also add the half body augmentation for COCO. The image size is $256\times 256$. 

In learning the representation, we use the AdamW~\cite{Loshchilov2019adamw} optimizer with the base learning rate set to $1e$-$2$ and weight decay to $0.15$, respectively. We warm up the learning rate for 500 iterations and drop the learning rate according to the cosine schedule. The batch size is $512$. We train $50$ epochs for 2D pose and $20$ epochs for 3D pose.

\subsection{Results on COCO, MPII and H36M}
\renewcommand{\arraystretch}{1.3}
\begin{table*}[t]
		\caption{Results on the COCO test-dev2017 and val2017 sets. The best results in the cited papers are reported. We set the batch size to $32$ when testing the speed of all models on a single V100 GPU. Since the official pre-trained model of Swin~\cite{liu2022swinv2} use square windows, we directly adopt the square input size to avoid domain gaps. While our input size seems larger than the competitors (\eg $256 \times 256$ vs. $256 \times 192$), the number of valid pixels is almost the same because the additional regions are mostly padded meaningless pixels.
		}
		\vspace{-0.25cm}
		\centering\setlength{\tabcolsep}{7.91pt}
		\label{table:coco}
		\footnotesize
		\begin{tabular}{l|cc|cc|ccc|ccc}
			\bottomrule
		    \multirow{2}{*}{Method} & \multirow{2}{*}{Backbone} & \multirow{2}{*}{Input size} & \multirow{2}{*}{GFLOPs $\downarrow$} & \multirow{2}{*}{Speed (fps) $\uparrow$} & \multicolumn{3}{c|}{COCO test-dev2017 $\uparrow$} & \multicolumn{3}{c}{COCO val2017 $\uparrow$}  \\
		    \cline{6-11}
			 & & & & & $\operatorname{AP}$ &$\operatorname{AP}^{50}$ &$\operatorname{AP}^{75}$ & $\operatorname{AP}$ &$\operatorname{AP}^{50}$ &$\operatorname{AP}^{75}$ \\
			\hline
			SimBa.~\cite{xiao18simpleb} & ResNet-152 & $384\times 288$ & $28.7$ & $76.3$ & $73.7$ & $91.9$ & $81.1$ & $74.3$ & $89.6$ & $81.1$\\
			PRTR~\cite{li2021prtr} & HRNet-W32 & $384\times 288$ & $21.6$ & $87.0$ & $71.7$ & $90.6$ & $79.6$ & $73.1$ & $89.4$ & $79.8$\\
			TransPose~\cite{yang2021transpose} & HRNet-W48 & $256\times 192$ & $21.8$ & $56.7$ & $75.0$ & $92.2$ & $82.3$ & $75.8$ & $90.1$ & $82.1$\\
			TokenPose~\cite{li2021tokenpose} & HRNet-W48 & $256\times 192$ & $22.1$ & $52.9$ & $75.9$ & $92.3$ & $83.4$ & $75.8$ & $90.3$ & $82.5$\\
			HRNet~\cite{wang19hrnetpami,sun2019hrnet} & HRNet-W48 & $384\times 288$ & $35.5$ & $75.5$ & $75.5$ & $92.7$ & $83.3$ & $76.3$ & $90.8$ & $82.9$\\
			DARK~\cite{zhang20dark} & HRNet-W48 & $384\times 288$ & $35.5$ & $62.1$ & $76.2$ & $92.5$ & $83.6$ & $76.8$ & $90.6$ & $83.2$\\
			UDP~\cite{huang20udp} & HRNet-W48 & $384\times 288$ & $35.5$ & $67.9$ & $76.5$ & $92.7$ & $84.0$ & $77.8$ & $92.0$ & $84.3$\\
			SimCC~\cite{li2021simcc} & HRNet-W48 & $384\times 288$ & $32.9$ & $71.4$ & $76.0$ & $92.4$ & $83.5$ & $76.9$ & $90.9$ & $83.2$ \\
			HRFormer~\cite{yuan21hrformer} & HRFormer-B & $384\times 288$ & $29.1$ & $25.2$ &  $76.2$ & $92.7$ & $83.8$ & $77.2$ & $91.0$ & $83.6$ \\
			ViTPose~\cite{xu2022vitpose} & ViT-Base & $256\times 192$ & $17.9$ & $113.5$ & $75.1$ & $92.5$ & $83.1$ & $75.8$ & $90.7$ & $83.2$ \\
			ViTPose~\cite{xu2022vitpose} & ViT-Large & $256\times 192$ & $59.8$ & $40.5$ & $77.3$ & $93.1$ & $85.3$ & $78.3$ & $91.4$ & $85.2$  \\
			ViTPose~\cite{xu2022vitpose} & ViT-Huge & $256\times 192$ & $122.9$ & $21.8$ & $78.1$ & $93.3$ & $85.7$ & $79.1$ & $91.6$ & $85.7$ \\
			SimBa.~\cite{xiao18simpleb} & Swin-Base & $256\times 256$ & $16.6$ & $74.4$ & $75.4$ & $93.0$ & $84.1$ & $76.6$ & $91.4$ & $84.3$ \\
			\hline
			Our approach & Swin-Base & $256\times 256$ & $15.2$ & $115.1$ & $76.5$ & $92.5$ & $84.7$ & $77.7$ & $91.2$ & $84.7$ \\
			Our approach & Swin-Large & $256\times 256$ & $34.1$ & $76.4$ & $77.4$ & $92.9$ & $85.2$ & $78.3$ & $91.4$ & $85.3$ \\
			Our approach & Swin-Huge & $256\times 256$& $118.2$ & $31.7$ & $78.3$ & $92.9$ & $85.9$ &  $79.3$ & $91.5$ & $85.9$\\
			\toprule
		\end{tabular}
		\vspace{-.3cm}
	\end{table*}

\paragraph{COCO.}~\Cref{table:coco} shows the results of the state-of-the-art top-down pose estimation methods on COCO~\cite{lin14coco} test-dev2017 and COCO val2017 sets, respectively. For our method, we provide three models of different sizes. We can see that they achieve better or comparable accuracy as the other methods. For example, our smallest model with Swin-Base outperforms the previous dominant heatmap-based methods including HRNet~\cite{sun2019hrnet}, HRFormer~\cite{yuan21hrformer}, and TokenPose~\cite{li2021tokenpose} with much faster inference speed. Similarly, our largest model also achieves better results than the state-of-the-art ViTPose (huge) with $1.5x$ faster inference speed. The fast inference speed is mainly due to the fact that our method does not require any expensive post-processing.

\paragraph{MPII.}
The results on the MPII validation set are shown in~\Cref{table:mpii}. The image size is set to be $256 \times 256$ for all methods. Our approach significantly surpasses the other methods. Our approach gets better performance mainly for the joints on the lower body which are easier to be occluded by other objects. Compared to the other classification-based method SimCC~\cite{li2021simcc}, our method achieves an improvement of $2.5$ under the metric of $\operatorname{PCKh@0.5}$. 

\renewcommand{\arraystretch}{1.3}
\begin{table}[t]
		\caption{Results on the MPII~\cite{Andriluka14mpii} val set (PCKh@0.5). }
		\vspace{-0.3cm}
		\centering\setlength{\tabcolsep}{3.10pt}
		\label{table:mpii}
		\footnotesize
		\begin{tabular}{l|ccccccc|c}
			\bottomrule
			Method & Hea. & Sho. & Elb. & Wri. & Hip. & Kne. & Ank. & Mean\\
			\hline
			SimBa.~\cite{xiao18simpleb}  & $97.0$ & $95.6$ & $90.0$ & $86.2$ & $89.7$ & $86.9$ & $82.9$ & $90.2$ \\
            PRTR~\cite{li2021prtr} & $97.3$ & $96.0$ & $90.6$ & $84.5$ & $89.7$ & $85.5$ & $79.0$ & $89.5$\\
			HRNet~\cite{wang20hrnet,sun2019hrnet} & $97.1$ & $95.9$ & $90.3$ & $86.4$ & $89.1$ & $87.1$ & $83.3$ & $90.3$ \\
			DARK~\cite{zhang20dark} & $97.2$ & $95.9$ & $91.2$ & $86.7$ & $89.7$ & $86.7$ & $84.0$ & $90.6$ \\
			TokenPose~\cite{li2021tokenpose} & $97.1$ & $95.9$ & $90.4$ & $86.0$ & $89.3$ & $87.1$ & $82.5$ & $90.2$ \\
			SimCC~\cite{li2021simcc} & $97.2$ & $96.0$ & $90.4$ & $85.6$ & $89.5$ & $85.8$ & $81.8$ & $90.0$ \\
			\hline
			Our (Swin-Base) & $97.5$ & $97.2$ & $92.8$ & $88.4$ & $92.4$ & $89.6$ & $87.1$ & $92.5$\\
			\toprule
		\end{tabular}
		\vspace{-.3cm}
	\end{table}
\renewcommand{\arraystretch}{1.4}
\begin{table}[h]
	\caption{3D pose estimation results on the Human3.6M~\cite{Ionescu14human36m} dataset. `*' means using extra 2D MPII~\cite{Andriluka14mpii} dataset for training. We report the MPJPE metric (mm). We only compare to the static image-based methods in the table.}
	\vspace{-0.3cm}
	\centering\setlength{\tabcolsep}{0.84pt}
	\label{table:h36m}
	\footnotesize
      \begin{tabular}{cccccccccccc|cc}
        \bottomrule
        \rotatebox{85}{Sharma \etal~\cite{sharma19gene}} &
        \rotatebox{85}{Zhao \etal~\cite{zhao2019semantic}} & \rotatebox{85}{Martinez \etal~\cite{Martinez17sim}} &
        \rotatebox{85}{Moon \etal~\cite{moon19topdown}} &
        \rotatebox{85}{Liu \etal~\cite{liu20weights}} & \rotatebox{85}{Xu and Takano~\cite{xu21hour}} & \rotatebox{85}{Li \etal~\cite{li20tagnet}} &\rotatebox{85}{Gong \etal~\cite{gong21poseaug}} &\rotatebox{85}{Zeng \etal~\cite{zeng20srnet}} & \rotatebox{85}{*Sun \etal~\cite{sun2018integral}} &\rotatebox{85}{Zou and Tang ~\cite{Zou21MGCN}}  & \rotatebox{85}{*Li \etal~\cite{li2021rle}} &\rotatebox{85}{Ours (Swin-Base)} &\rotatebox{85}{Ours (Swin-Huge)}\\
        \hline
        $58.0$ & $58.0$ & $57.6$ & $54.4$ & $52.4$ & $51.9$ & $50.9$ & $50.2$ & $49.9$ & $49.6$ & $49.4$ & $48.6$  & $50.8$ & $47.8$\\
        \toprule
      \end{tabular}
		\vspace{-.3cm}
\end{table}

\paragraph{H36M.}

It is straightforward to apply the PCT representation to 3D pose estimation. We first learn the encoder, the codebook and the decoder on the 3D poses. Then we train a classification head for 3D pose estimation. For simplicity, we directly use the backbone used in 2D pose estimation without re-training. The results are shown in~\Cref{table:h36m}.  Our approach achieves a smaller error than the state-of-the-art monocular image-based methods. The results show that PCT is general and applies to both 2D and 3D poses.

			
	
\subsection{Results on CrowdPose, OCHuman, SyncOCC}
	\renewcommand{\arraystretch}{1.25}
	\begin{table*}[t]
		\caption{The results of the state-of-the-art methods on the occlusion datasets. The numbers of the competitors are obtained by running their official models using the MMPose~\cite{mmpose2020} framework. The metrics are computed only on the occluded joints that overlap with the COCO annotated joints. The GT bounding box is used. `OC' denotes the OCHuman~\cite{zhang19ochuman} dataset.}
		\vspace{-.3cm}
		\centering\setlength{\tabcolsep}{3.85pt}
		\label{table:occlusionsota}
		\footnotesize
		\begin{tabular}{l|ccc|ccccc}
			\bottomrule
		    \multirow{2}{*}{Method}  &  \multirow{2}{*}{Backbone} &  \multirow{2}{*}{Input size} &  \multirow{2}{*}{Speed (fps) $\uparrow$\textbf{}} & \multicolumn{5}{c}{2D Occluded Pose Estimation ($\operatorname{AP}^{OC} \uparrow$)} \\
		    \cline{5-9}
			 & & & &  OC-val~\cite{zhang19ochuman} & OC-test~\cite{zhang19ochuman} & CrowdPose~\cite{li2018crowdpose} & SyncOCC~\cite{zhang2020syncocc} & SyncOCC-H~\cite{zhang2020syncocc}\\
			\hline
			HRNet~\cite{wang19hrnetpami,sun2019hrnet} & HRNet-W48 & $384\times 288$ & $75.5$ & $38.1$ & $38.1$ & $74.5$  & $90.8$ & $73.0$\\
			DARK~\cite{zhang20dark} & HRNet-W48 & $384\times 288$ & $62.1$ & $38.6$ & $39.2$ & $74.9$ & $91.2$ & $73.8$\\
			UDP~\cite{huang20udp} & HRNet-W48 & $384\times 288$ & $67.9$ & $38.6$ & $38.8$ & $75.0$ & $90.8$ & $73.0$\\
			HRFormer~\cite{yuan21hrformer} & HRFormer-B & $384\times 288$ & $25.2$ & $40.5$ & $40.3$ & $72.4$ & $91.9$ & $75.7$ \\
			Poseur~\cite{mao2022poseur} & HRFormer-B & $384\times 288$ & $25.8$ & $44.4$ & $45.6$ &$73.9$ &$93.1$ & $78.5$\\
			ViTPose~\cite{xu2022vitpose}  & ViT-Huge & $256\times 192$ & $21.8$ & $46.7$ &$45.8$ &$74.7$ &$92.3$ & $77.4$\\
			SimBa.~\cite{xiao18simpleb} & Swin-Base & $256\times 256$ & $74.4$ & $40.1$ & $39.8$ &$71.6$ &$90.7$ & $72.4$\\
			\hline
			Our approach & Swin-Base & $256\times 256$ & $115.1$ & $45.6$ &$44.5$ & $73.9$ & $93.0$  & $78.3$\\
			Our approach & Swin-Large & $256\times 256$ & $76.4$ & $47.2$ & $47.0$ & $76.8$ & $93.4$  & $78.9$\\
			Our approach & Swin-Huge & $256\times 256$ & $31.7$ & $\textbf{50.8}$ & $\textbf{49.6}$ & $\textbf{77.2}$ & $\textbf{94.0}$  & $\textbf{79.7}$\\
			\toprule
		\end{tabular}
		\vspace{-.32cm}
	\end{table*}
	
	\renewcommand{\arraystretch}{1.25}
	\begin{table}[t]
		\caption{Comparison of four pose representations in a completely fair setting. We conduct the experiments with the Swin-Base and input size $256\times 256$. The results are reported on the occluded joints that overlap with the COCO annotated joints. The GT bounding box is used. `OC' denotes the OCHuman~\cite{zhang19ochuman} dataset. }
		\vspace{-0.2cm}
		\centering\setlength{\tabcolsep}{1.9pt}
		\label{table:occcomp}
		\footnotesize
		\begin{tabular}{l|ccccc}
			\bottomrule
		    Method  & \ \ OC-val\ \ & \ \ OC-test\ \ & CrowdPose & SyncOCC & SyncOCC-H \\
			\hline
			Heatmaps & $40.1$ & $39.8$ &$71.6$ &$90.7$ & $72.4$ \\
			Discrete Bins  & $40.5$ &$39.9$ &$71.9$ &$91.1$ & $73.6$\\
			Coordinates  & $41.5$ & $41.5$  & $72.7$ & $91.9$  & $75.7$ \\
			Our PCT & $\textbf{45.6}$ &$\textbf{44.5}$ & $\textbf{73.9}$ & $\textbf{93.0}$  & $\textbf{78.3}$ \\
			\toprule
		\end{tabular}
		\vspace{-.08cm}
	\end{table}
	We evaluate how our method performs in severe occlusions. The results on the four occlusion datasets are shown in~\Cref{table:occlusionsota}.
    We can see that our PCT based approach significantly outperforms the other methods. ~\Cref{fig:qualitive} shows some examples. There are several interesting observations. First, when a large portion of the human body is occluded, our method can predict a reasonable configuration for the occluded joints that is in harmony with the visible joints although there are no supporting visual features. This validates the strong context modeling capability of our method. Second, when a small portion is occluded, our method can predict accurate positions based on the visual features in the neighborhood. For example, in the fourth example of the first row, the ankle joint of the rightmost person is correctly predicted based on the visual features of the legs. Third, it also shows stronger capability to resolve the ambiguities of other distracting persons.
 
 \begin{table}[t]
		\centering
        \setlength{\tabcolsep}{7.0pt}
        \footnotesize
            \renewcommand{\arraystretch}{1.2}
			\newcommand{\tabincell}[2]{\begin{tabular}{@{}#1@{}}#2\end{tabular}}
			\caption{Ablation study of four main components: Compo (compositional design), MJM (masked joint modeling), IG (image guidance), and RecLoss (auxiliary pose reconstruction Loss). We report the $\operatorname{AP}^{V}$ for reconstructed poses, $\operatorname{AP}^{P}$ for predicted poses on the COCO val2017 set, and $\operatorname{AP}^{OC}$ on the SyncOCC test set.
			All results are obtained with the backbone Swin-Base and input size $256\times 256$.
			}
		    \vspace{-0.2cm}
			\begin{tabular}{ c  c c c  | c c  c }
			\bottomrule
			Compo & MJM &  IG & RecLoss & $\operatorname{AP}^{V}$ & $\operatorname{AP}^{P}$ & $\operatorname{AP}^{OC}$\\
				\hline
				& &  & & $33.1$ & $16.2$ & $56.8$\\ 
				\checkmark &   &   &   & $98.9$ & $65.5$ & $88.3$\\
				\checkmark & \checkmark  &  &   & $99.0$ & $72.7$ & $91.2$\\
				\checkmark & \checkmark  & \checkmark  &  & $99.0$ & $75.1$ & $92.8$\\
				\checkmark & \checkmark  & \checkmark  & \checkmark  & $99.0$ & $77.4$ & $93.1$\\
			\toprule
			\end{tabular}
			\label{tab:ablation-study-final}
		    \vspace{-0.34cm}
	\end{table}
	
    We also compare the four representations including the coordinates, heatmaps, discrete bins, and PCT in a completely fair setting. The results are shown in
    \Cref{table:occcomp}. We can see that PCT achieves much better results than the dominant heatmap representation, leading by about $5.0$ AP on OCHuman, $2.3$ AP on SyncOCC, and $5.9$ AP on the more challenging SyncOCC hard set.

\subsection{Empirical analysis}

 \begin{figure}
 \footnotesize
    \centering
    \includegraphics[width = 1.0\linewidth]{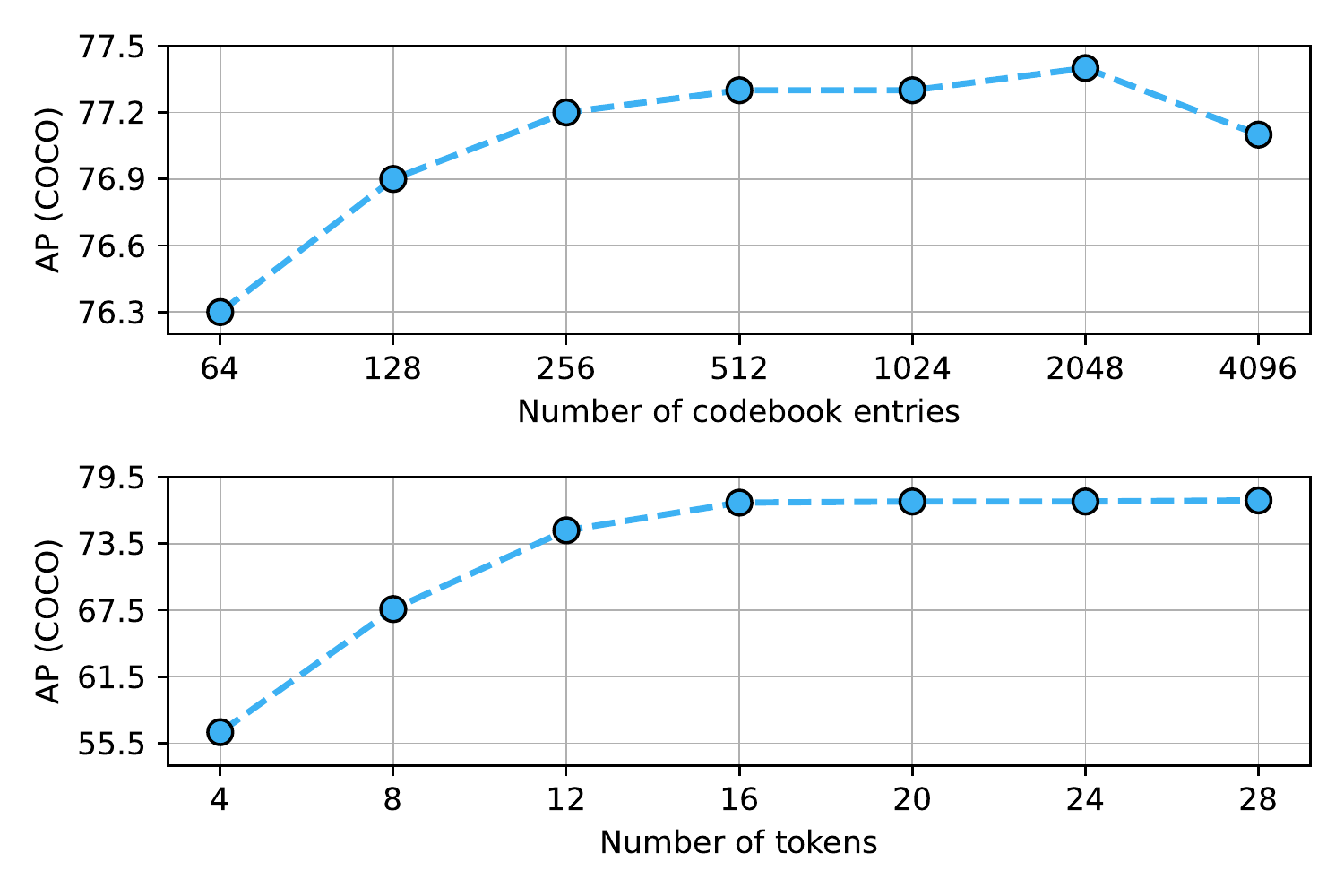}
    \vspace{-2.0em}
    \caption{Impact of the number of codebook entries and the number of tokens, respectively. The results are obtained by the model using the Swin-Base backbone trained for 150 epochs on the COCO val2017 dataset. 
    }
    \vspace{-0.35cm}
    \label{fig:layerwisequalitive}
\end{figure}

\begin{figure*}
	\footnotesize
	\centering
    \includegraphics[height=0.176\linewidth, trim={10 0 10 5},clip]{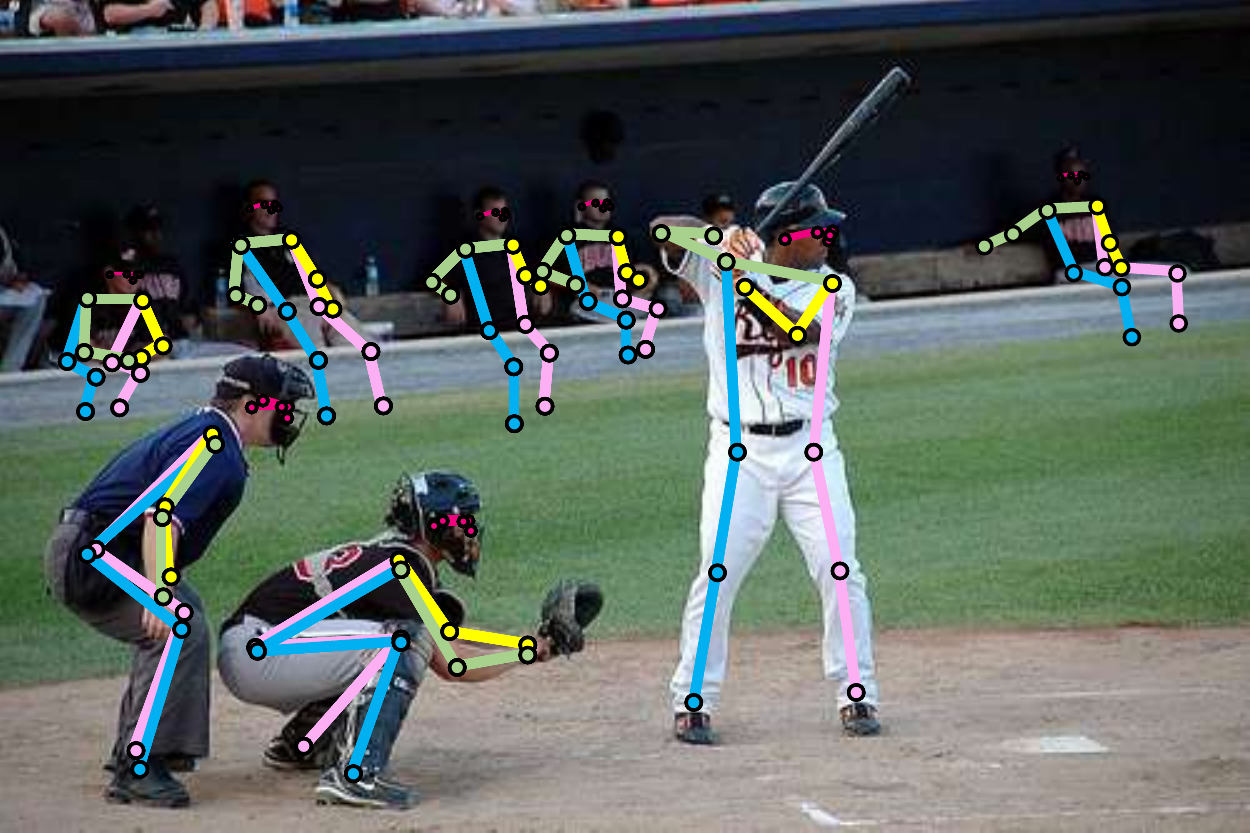}~
    \includegraphics[height=0.176\linewidth, trim={0 0 156 20},clip]{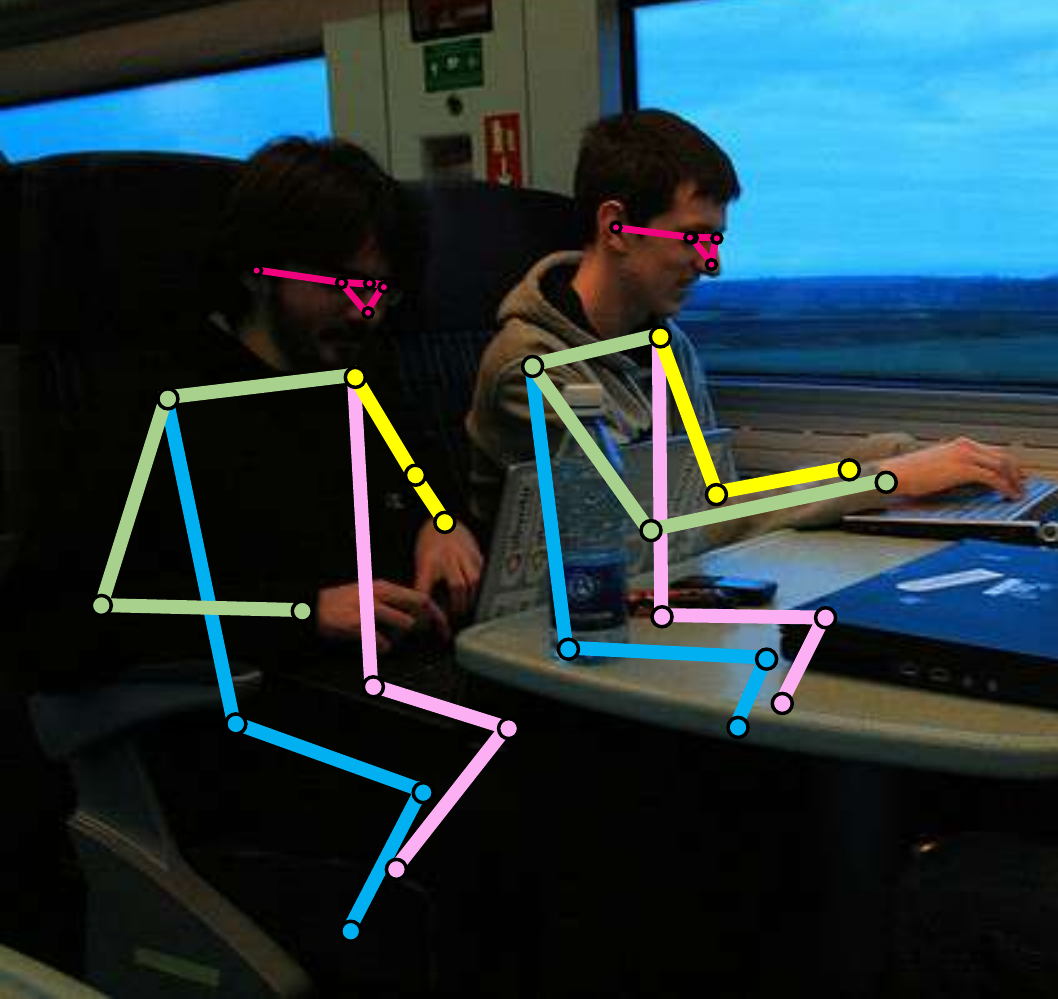}~
    \includegraphics[height=0.176\linewidth, trim={20 110 180 50},clip]{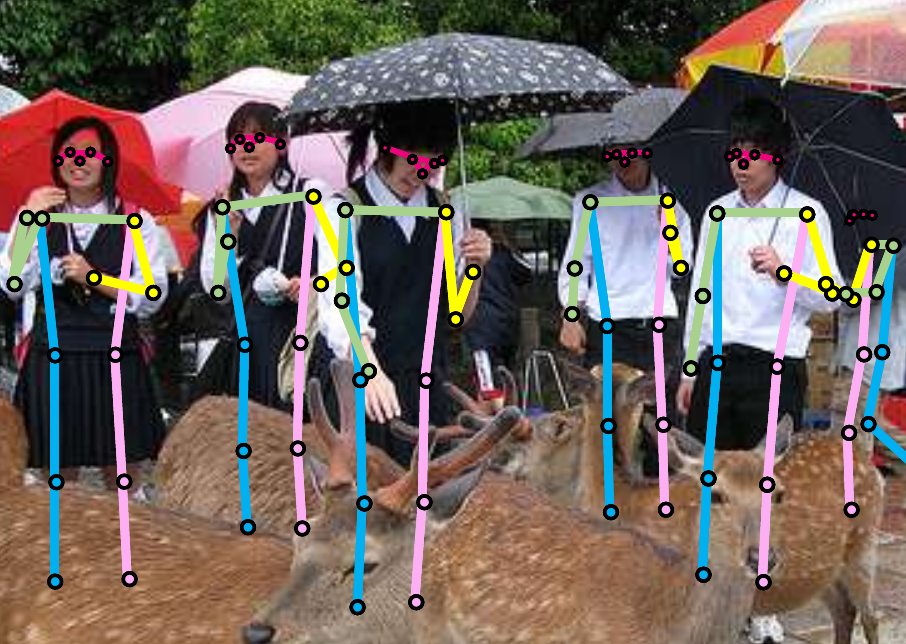}~
    \includegraphics[height=0.176\linewidth, trim={40 77 0 22},clip]{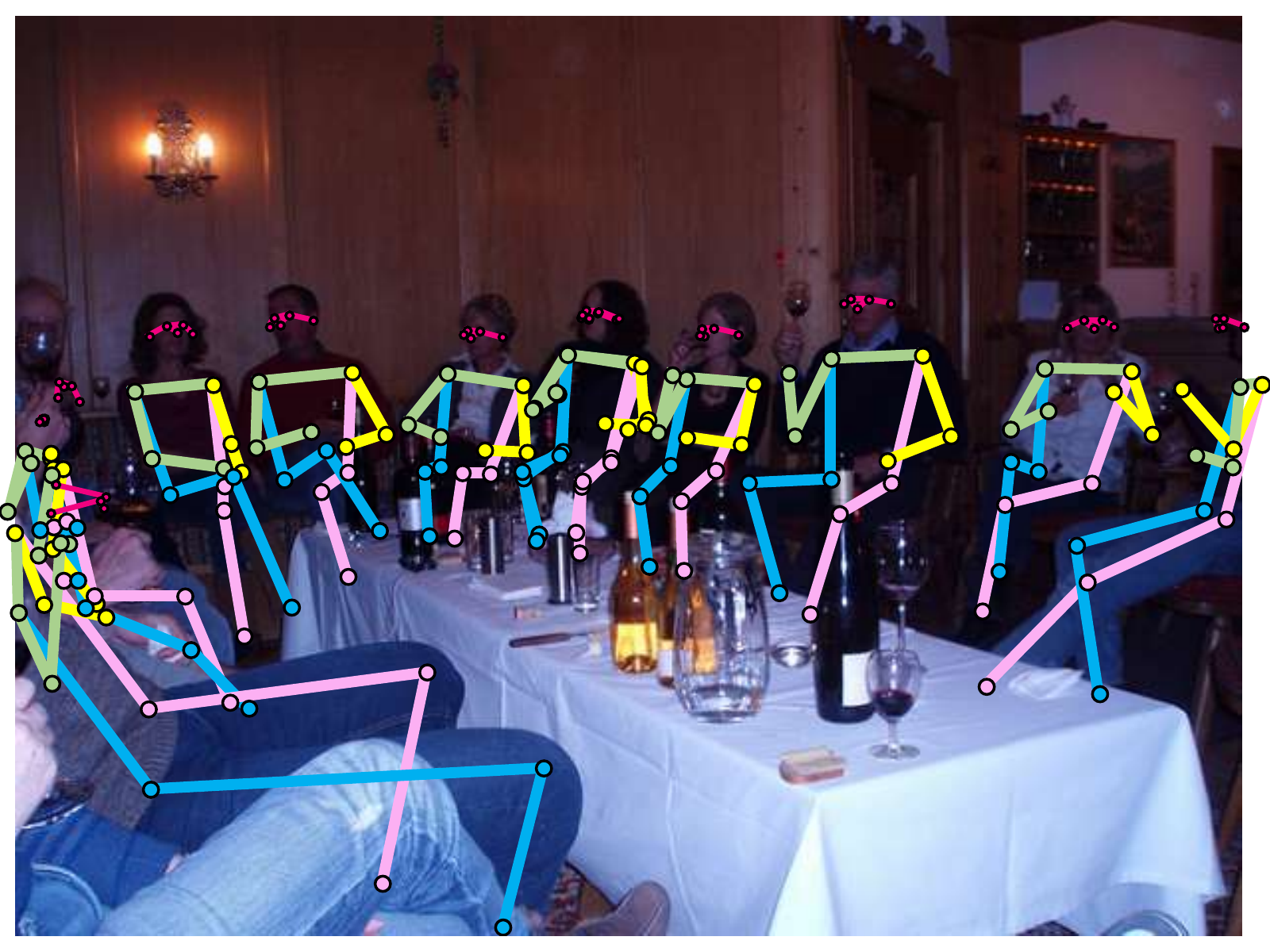}~\\
    \vspace{0.06cm}
    \includegraphics[height=0.176\linewidth]{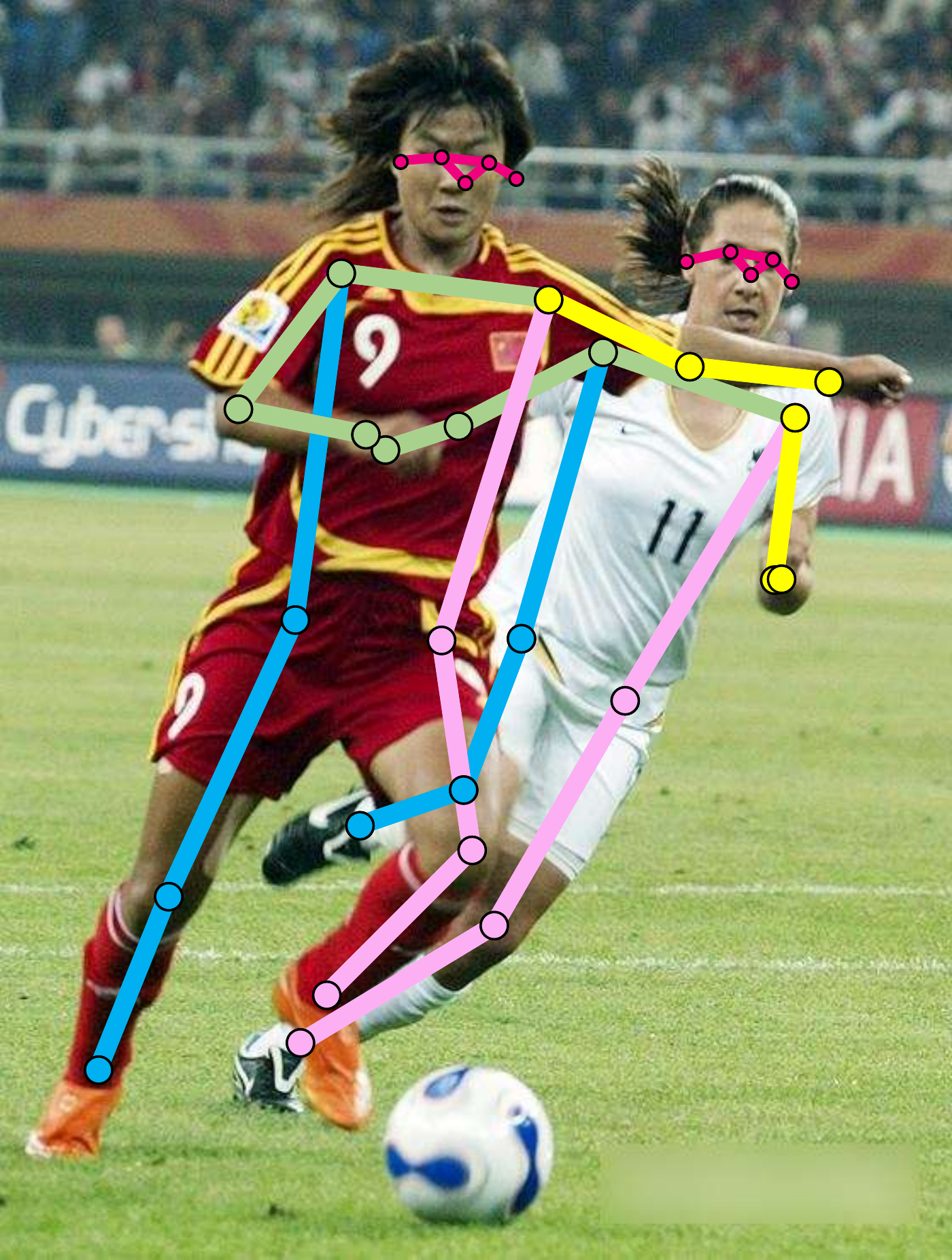}~
    \includegraphics[height=0.176\linewidth, trim={140 30 60 40},clip]{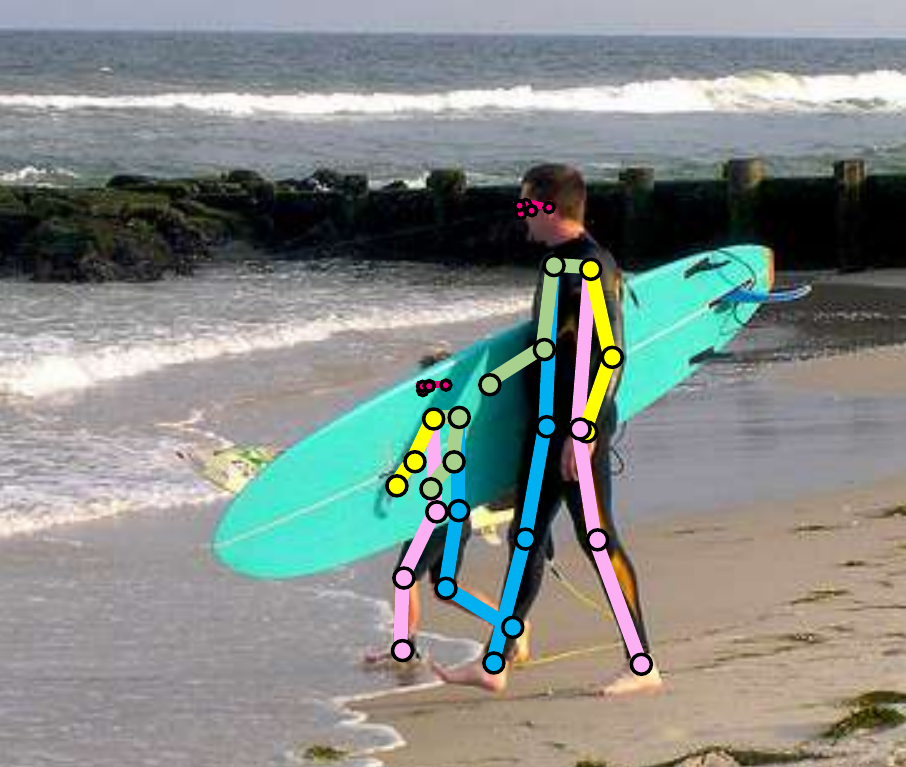}~
    \includegraphics[height=0.176\linewidth]{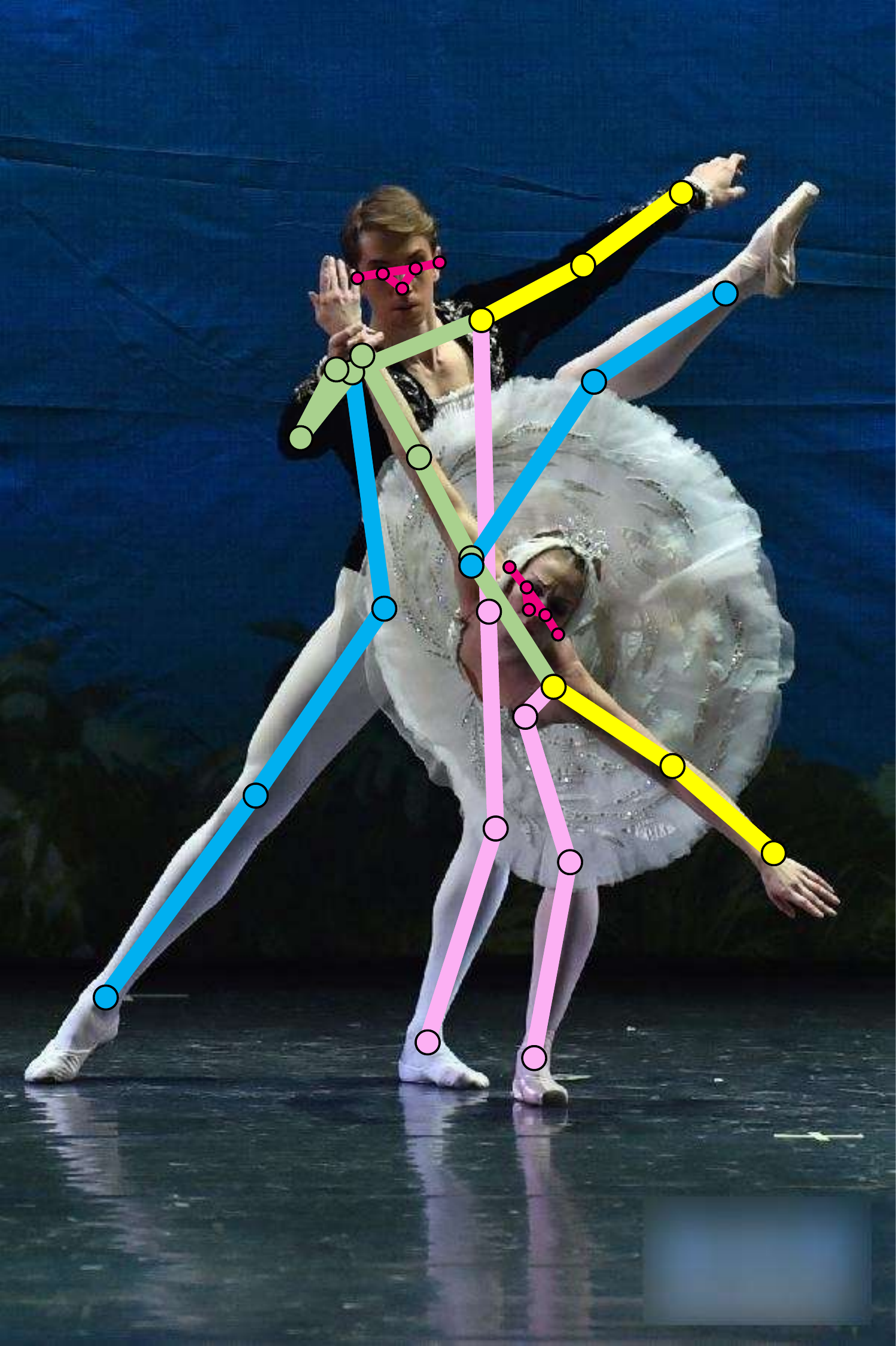}~
    \includegraphics[height=0.176\linewidth]{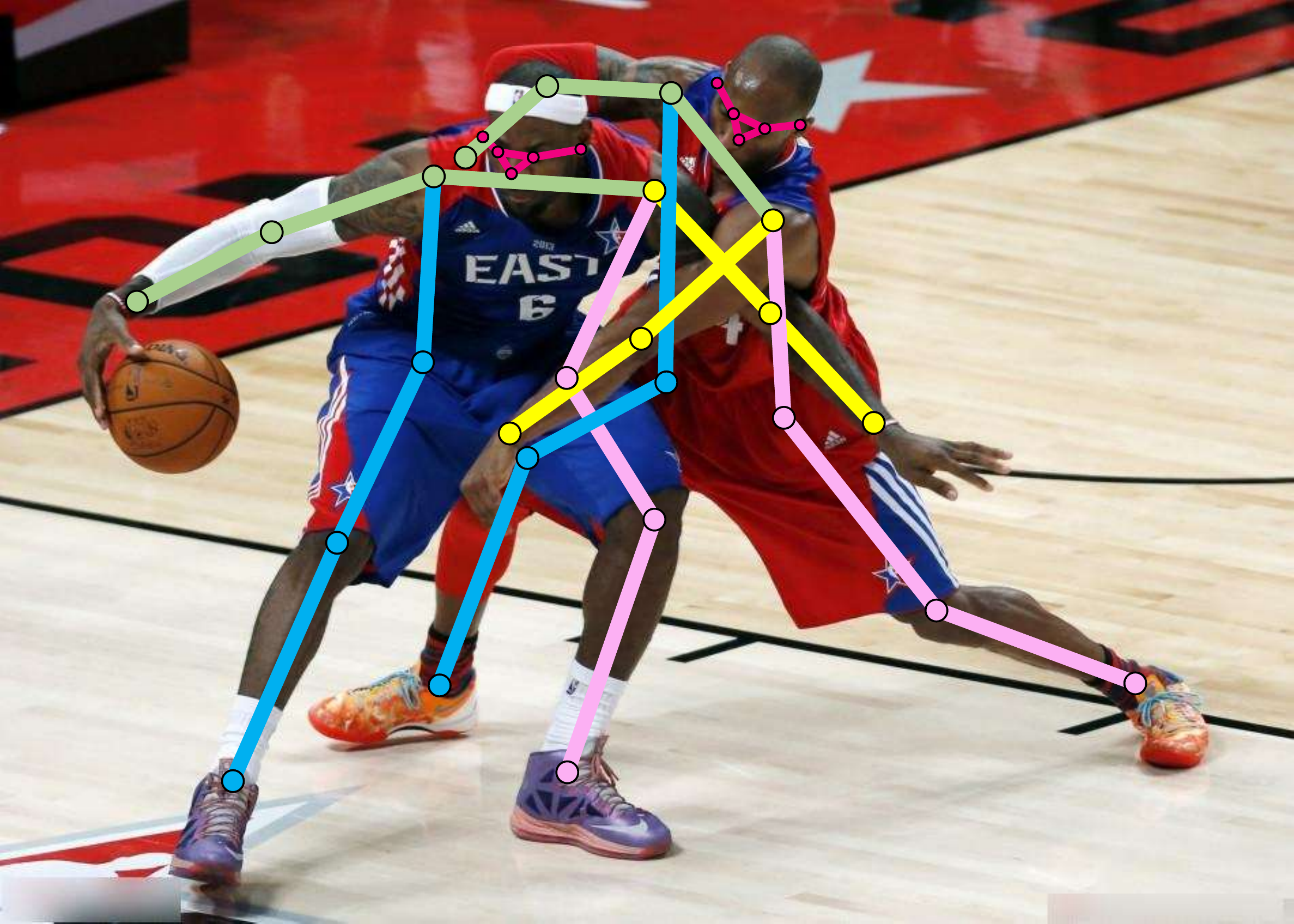}~
    \includegraphics[height=0.176\linewidth, trim={50 0 100 0},clip]{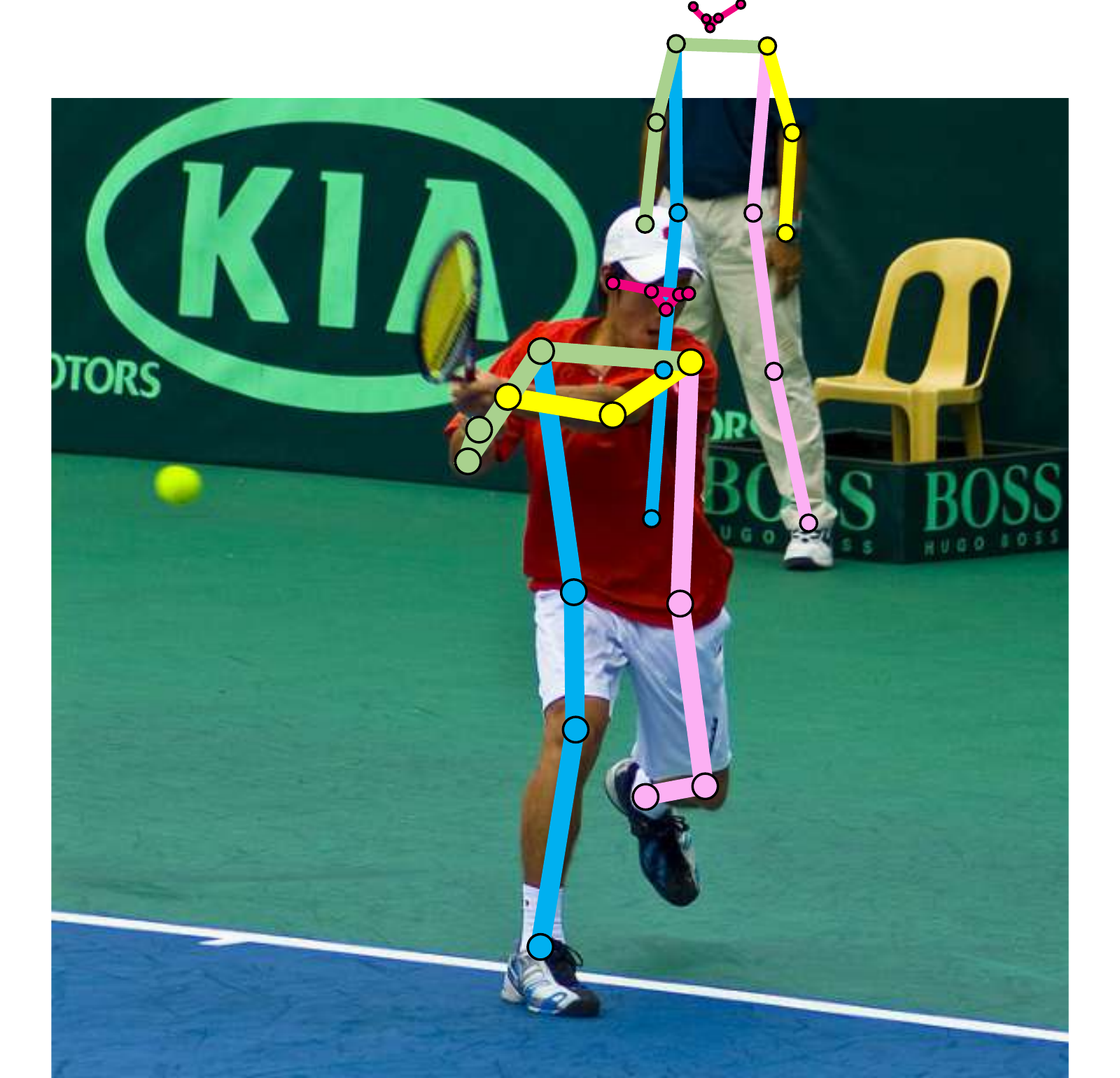}~
    \includegraphics[height=0.176\linewidth, trim={255 0 255 0},clip]{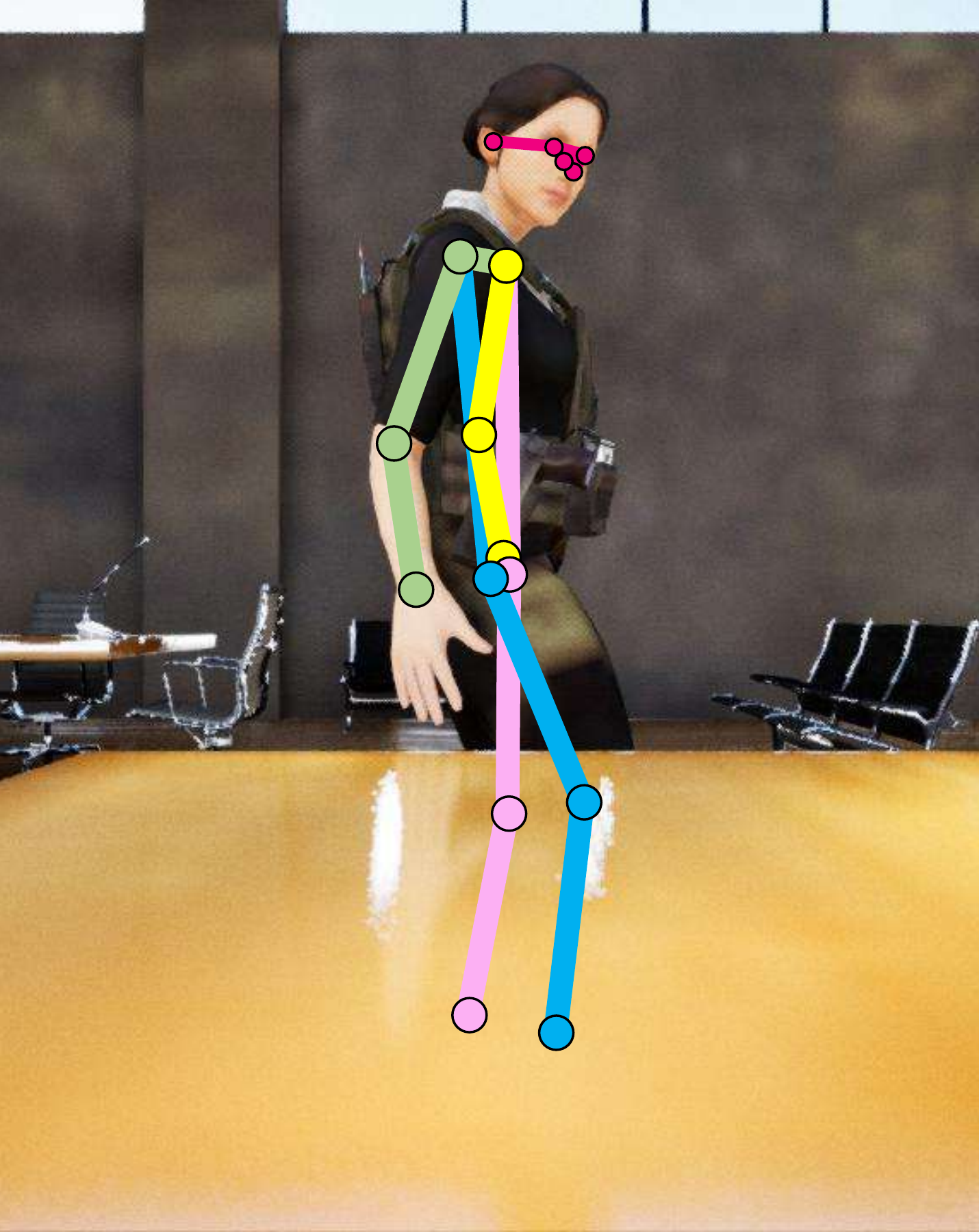}~\\
    \vspace{0.06cm}
    \includegraphics[height=0.176\linewidth, trim={255 0 254 0},clip]{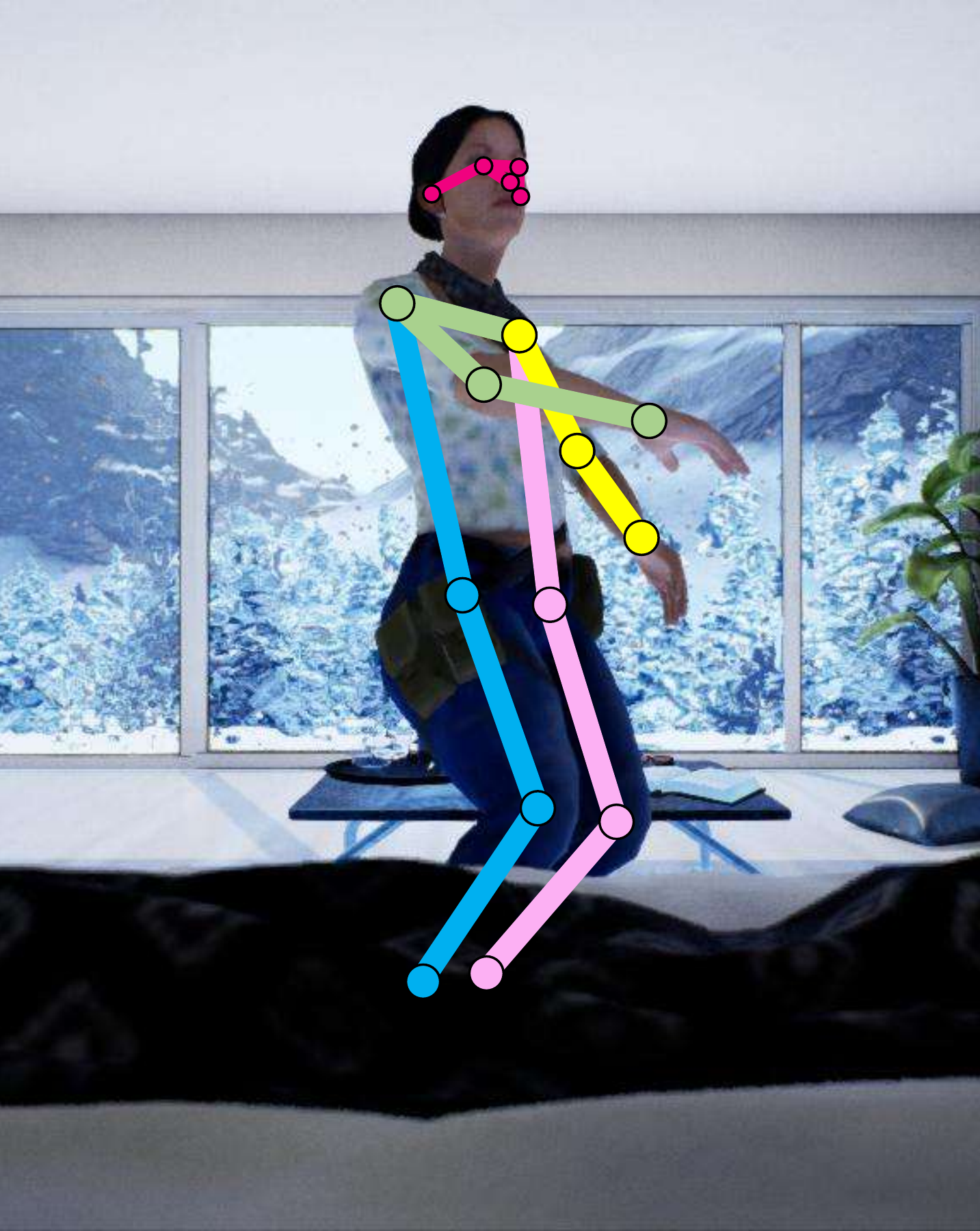}~
    \includegraphics[height=0.176\linewidth, trim={25 30 25 10},clip]{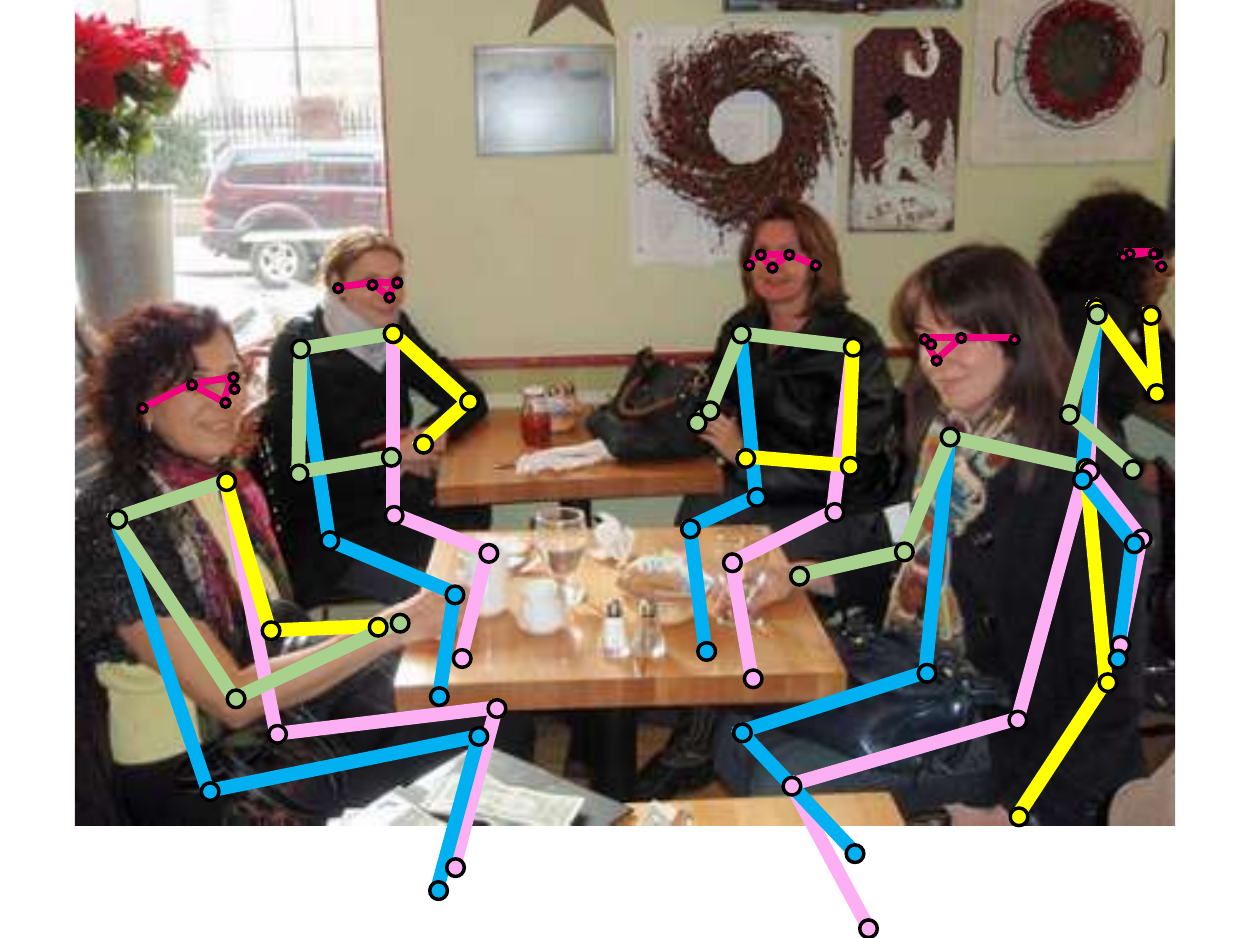}~
    \includegraphics[height=0.176\linewidth, trim={170 20 30 110},clip]{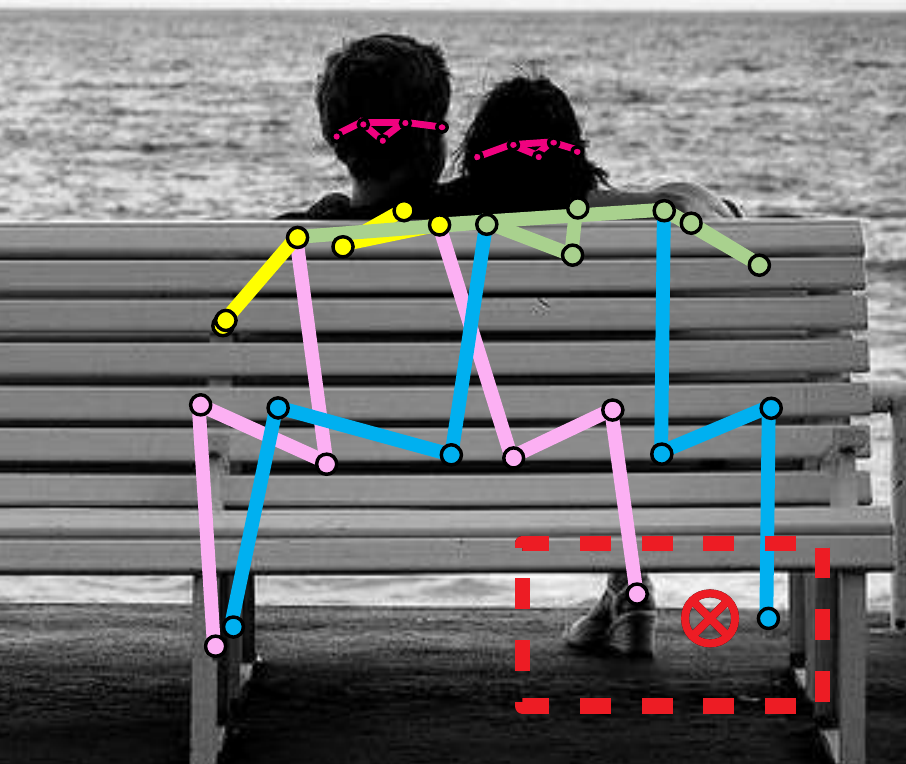}~
    \includegraphics[height=0.176\linewidth, trim={10 0 250 150},clip]{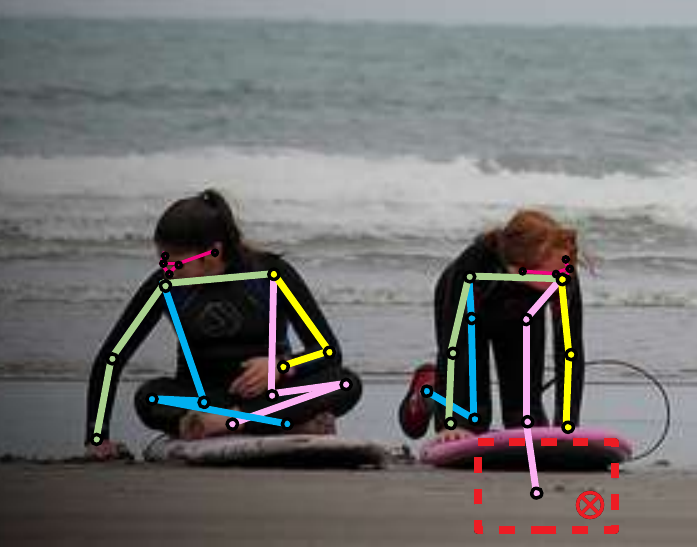}~
    \includegraphics[height=0.176\linewidth, trim={80 10 180 110},clip]{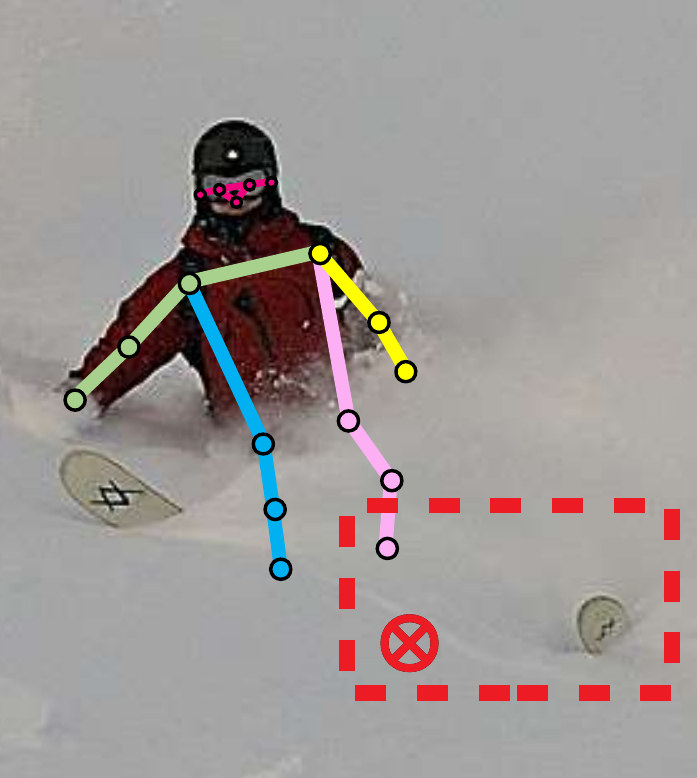}~\\
    \vspace{-0.1cm}
	\caption{
    Qualitative results of our approach with Swin-Base backbone. The images are obtained from OChuman test set, COCO val2017 set, CrowdPose test set, and SyncOCC.}
	\label{fig:qualitive}
	\vspace{-0.3cm}
\end{figure*}

\paragraph{Ablation study.}
We ablate the main components of PCT that we think are important. It includes the Compositional design (Compo), Masked Joints Modeling (MJM), Image Guidance (IG), and auxiliary Pose Reconstruction Loss (RecLoss). All experiments are conducted on the COCO val set and the SyncOCC set, using the Swin-Base backbone trained for $150$ epochs.

The first baseline discards the compositional design and learns a codebook for each joint without interactions between the joints. As can be seen in~\Cref{tab:ablation-study-final}, $\operatorname{AP}^{V}$ is only $33.1\%$ meaning that the codebook cannot even reconstruct the poses accurately.  This is because we need a significantly larger codebook without the compositional design. As a result, the pose estimation accuracy $\operatorname{AP}^{P}$ in the downstream task is only $16.2\%$. Adding the compositional design directly improves $\operatorname{AP}^{V}$ to $98.9\%$. Adding MJM improves $\operatorname{AP}^{P}$ significantly from $65.5\%$ to $72.7\%$. Our understanding is that MJM can drive the model to learn meaningful sub-structures (tokens) to help detect masked joints. IG and RecLoss also improve the results. 

\paragraph{Token number.}
Increasing the number of tokens $M$ will enlarge the representation space exponentially. The results are shown in ~\Cref{fig:layerwisequalitive}. We can see that increasing $M$ from $4$ to $16$ notably improves the AP on the COCO dataset. Further increasing $M$ brings little improvement. We find this is because the newly added tokens become redundant and have a large overlap with the existing ones. However, the results are barely affected by the redundant tokens which make the approach robust to the parameter. 

\paragraph{Codebook size.}
Increasing the number of entries $V$ in the codebook decreases the quantization error. However, it also increases the classification difficulty as the number of categories becomes larger. The results are shown in~\Cref{fig:layerwisequalitive}. Setting this number between $256$ and $2048$ gives satisfactory results. Again, the model is not very sensitive to this parameter.

\paragraph{Qualitative results.}~\Cref{fig:qualitive} shows some pose estimation results. We can see that it handles occlusion in a reasonable way. When a human body is occluded by a large region where even people are not completely sure about the exact pose, our method can predict a reasonable pose although it may be different from the GT pose. Note that they are not cherry-picked results. The last three examples show the failure cases. For the two people on the chair example, it is probable that the right ankle joint should be somewhere occluded by the chair. Similarly, for the person skating example, the ankle joints should be near the skateboard. The results suggest that leveraging objects as the context may further improve the estimation results. 

\section{Conclusion}
In this work, we introduce a structured representation PCT to the human pose estimation community, which models the dependency between the body joints and automatically learns the sub-structures of the human pose. We also present a very simple pose estimation pipeline on top of the PCT representation, which does not need any complicated post-processing. It achieves better or comparable results as the state-of-the-art methods on five benchmarks. The discrete representation also paves the way for interacting with other discrete modalities such as text and speech. 

\noindent{\textbf{Future work.}} It will be interesting to further reduce the ambiguities in pose estimation by exploring other cues under the discrete representation. For example, as mentioned in the qualitative study, we can model the context from the environments such as the surrounding objects.

\appendix

\begin{figure*}
	\footnotesize
	\centering
\includegraphics[width=0.8\textwidth, trim={0 470 10 10},clip]{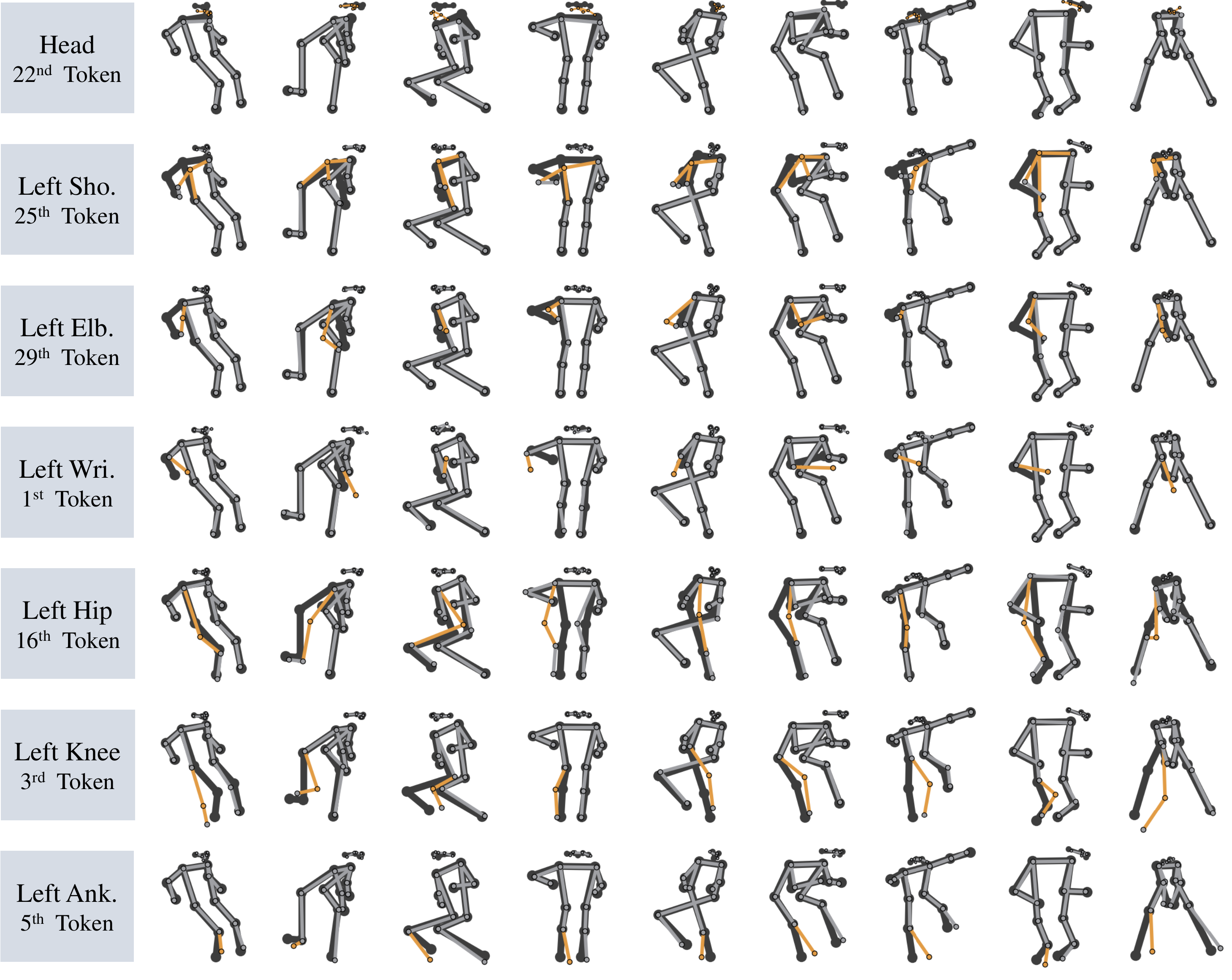}

	\caption{
	Each token is learned to represent a sub-structure. In each row, we show that if we change the stage of one token to different values, it consistently changes the same sub-structure highlighted by orange. The black poses are before changing.}
	\label{fig:substructure}
	\vspace{-0.25cm}
\end{figure*}

\section*{Appendix}

\subsection*{Results on the MPII Test Set}
We provide the results on the MPII~\cite{Andriluka14mpii} test set. ~\Cref{table:mpii} shows the results on the MPII test set. Our approach outperforms the other methods, even those that utilize extra training datasets or larger image sizes.

 \subsection*{Results on the H36M under occlusion}
 To evaluate the performance of PCT under different occlusion conditions, we artificially occlude the images in the h36m test set by either cropping or masking them. \Cref{table:h36mocc} reports the results of the models with and without PCT. It reveals that the advantages of PCT become more apparent as the level of occlusion increases.

\subsection*{More visual illustrations for the sub-structures.}
\label{sec:sub-structure}
~\Cref{fig:substructure} provides more examples of sub-structures represented by our compositional tokens. We use $34$ tokens to represent a human pose. We find that almost two tokens are responsible for a sub-structure consisting of a body joint and its related joints, one is for major changes, and the other is for minor jitters. We select some of them to show.

\renewcommand{\arraystretch}{1.28}
\begin{table}[t]
		\caption{Results on the MPII~\cite{Andriluka14mpii} test set (PCKh@0.5). '$^{\dagger}$' means using extra training datasets. '$^{\ddagger}$' means using larger image size.}
		\vspace{-0.3cm}
		\centering\setlength{\tabcolsep}{2.95pt}
		\label{table:mpii}
		\footnotesize
		\begin{tabular}{l|ccccccc|c}
			\bottomrule
			Method & Hea. & Sho. & Elb. & Wri. & Hip. & Kne. & Ank. & Mean\\
			\hline
			Xiao \etal~\cite{xiao18simpleb}  & $98.5$ & $96.6$ & $91.9$ & $87.6$ & $91.1$ & $88.1$ & $84.1$ & $91.5$ \\
			Tang \etal~\cite{tang18comp} & $98.4$ & $96.9$ & $92.6$ & $88.7$ & $91.8$ & $89.4$ & $86.2$ & $92.3$ \\
			Sun \etal~\cite{wang20hrnet,sun2019hrnet} & $98.6$ & $96.9$ & $92.8$ & $89.0$ & $91.5$ & $89.0$ & $85.7$ & $92.3$ \\
                Cai \etal~\cite{cai20rsn} & $98.5$ & $97.3$ & $93.9$ & $89.9$ & $92.0$ & $90.6$ & $86.8$ & $93.0$ \\
                Bulat \etal~\cite{Bulat20softgated}$^{\dagger}$ & $98.8$ & $97.5$ & $94.4$ & $91.2$ & $93.2$ & $92.2$ & $89.3$ & $94.1$ \\
                Bin \etal~\cite{bin20asda}$^{\ddagger}$ & $98.9$ & $97.6$ & $94.6$ & $91.2$ & $93.1$ & $92.7$ & $89.1$ & $94.1$ \\
			\hline
			Our (Swin-Base) & $98.7$ & $97.5$ & $94.2$ & $90.6$ & $92.9$ & $92.1$ & $88.7$ & $93.8$\\
			Our (Swin-Large) & $98.9$ & $97.8$ & $94.8$ & $91.1$ & $93.6$ & $93.0$ & $89.7$ & $94.3$\\
			\toprule
		\end{tabular}
		\vspace{-.1cm}
	\end{table}

\renewcommand{\arraystretch}{1.28}
\begin{table}[t]
    \caption{Results on the H36M~\cite{Ionescu14human36m} test set (MPJPE mm) under different occlusion conditions. }
    \vspace{-0.3cm}
    \centering\setlength{\tabcolsep}{9.7pt}
    \label{table:h36mocc}
    \footnotesize
    \begin{tabular}{l|ccccc}
        \bottomrule
        Mask Ratio & 0.0 & 0.2 & 0.4 & 0.6 & 0.8\\
        \hline
        w/o PCT  & $53.9$ & $66.6$ & $94.4$ & $157.6$ & $268.7$ \\
        PCT & $50.8$ & $63.4$ & $88.2$ & $145.5$ & $287.9$ \\
        \hline
        \hline
        Crop Ratio & 0.0 & 0.1 & 0.2 & 0.3 & 0.4\\
        \hline
        w/o PCT  & $53.9$ & $53.9$ & $54.8$ & $60.0$ & $84.1$ \\
        PCT & $50.8$ & $50.9$ & $51.2$ & $55.0$ & $74.8$ \\
        \toprule
    \end{tabular}
    \vspace{-.3cm}
\end{table}
{\small
\bibliographystyle{ieee_fullname}
\bibliography{posetoken}
}

\end{document}